# Chemical classification program synthesis using generative artificial intelligence

## Authors

Mungall, Christopher J.[1]*
Malik, Adnan[2]
Korn, Daniel R.[3]
Reese, Justin T.[1]
O'Boyle, Noel M.[2]
Hastings, Janna[4,5,6]

1. Division of Environmental Genomics and Systems Biology, Lawrence Berkeley National Laboratory, Berkeley, CA 94720, USA
2. European Molecular Biology Laboratory, European Bioinformatics Institute (EMBL-EBI), Wellcome Genome Campus, Hinxton, Cambridge CB10 1SD, United Kingdom
3. Department of Computer Science, The University of North Carolina at Chapel Hill College of Arts and Sciences, Chapel Hill, NC 27599, USA
4. School of Medicine, University of St. Gallen, St. Gallen, Switzerland
5. Institute for Implementation Science in Health Care, University of Zurich, Zurich, Switzerland
6. Swiss Institute of Bioinformatics, Lausanne, Switzerland

*Corresponding author

# Abstract

Accurately classifying chemical structures is essential for cheminformatics and bioinformatics, including tasks such as identifying bioactive compounds of interest, screening molecules for toxicity to humans, finding non-organic compounds with desirable material properties, or organizing large chemical libraries for drug discovery or environmental monitoring. However, manual classification is labor-intensive and difficult to scale to large chemical databases. Existing automated approaches either rely on manually constructed classification rules, or are deep learning methods that lack explainability.

This work presents an approach that uses generative artificial intelligence to automatically write *chemical classifier programs* for classes in the Chemical Entities of Biological Interest (ChEBI) database. These programs can be used for efficient deterministic run-time classification of SMILES structures, with natural language explanations. The programs themselves constitute an explainable computable ontological model of chemical class nomenclature, which we call the ChEBI Chemical Class Program Ontology (C3PO).

We validated our approach against the ChEBI database, and compared our results against deep learning models and a naive SMARTS pattern based classifier. C3PO outperforms the naive classifier, but does not reach the performance of state of the art deep learning methods. However, C3PO has a number of strengths that complement deep learning methods, including explainability and reduced data dependence. C3PO can be used alongside deep learning classifiers to provide an explanation of the classification, where both methods agree. The programs can be used as part of the ontology development process, and iteratively refined by expert human curators.

## Scientific Contribution

We demonstrate a novel knowledge distillation technique in which the classifiers are programs, leveraging the power of cheminformatics software libraries. We demonstrate applicability for classifying chemical structures, and for assisting curation of chemical databases.

# Introduction

## Scientific insight relies on chemical classification

Chemical databases such as PubChem[1] and ChEMBL[2] include hundreds of millions of structures, and the number of potential drug-like structures has been estimated at $10^{60}$[3]. Organizing, grouping, and classifying these structures into classes of chemical structures is an important task for many applications, particularly in the medical and biological sciences[4]. For example, medicinal chemists might want to scan genome databases to find pathways for producing terpenoids, a class of natural products derived from isoprene units with many applications. Drug developers might want to identify classes of molecules suitable for drug-target modeling. Environmental health scientists might want to explore the effects of exposure to phenol compounds, another large and useful class of chemicals. There are thousands of such potential classes, often organized into hierarchical systems. For example, terpenoids can be further classified based on the number of isoprene units into subclasses such as monoterpenoids, sequiterpenoids, and so on.

One of the main systems used for chemical structure classification in the life sciences is the Chemical Entities of Biological Interest (ChEBI) database[5], which contains the classification of almost 200,000 chemical structures according to thousands of predefined classes, arranged as an ontological network. ChEBI has proven to be a vital cheminformatics and bioinformatics resource, used in a variety of applications, databases, knowledge bases, and ontologies. For example, ChEBI is used by the Gene Ontology (GO) project for chemical classification[6] and metabolic pathway curation[7]. ChEBI is also used by metabolomics databases such as MetaboLights in order to standardize sample metabolite assignments[8].

A major bottleneck in ChEBI development is that classification has historically been largely manual, relying on expert knowledge and curation, which is inherently time-consuming and becomes increasingly challenging as the size of chemical databases grows (the number of structures in ChEBI is a fraction of those in larger databases). Even within the more modest scale of the ChEBI database itself, the manual classification approach often leads to incomplete or inconsistent classifications. There is therefore an urgent need to provide explainable automated methods to curators to assist with the classification task.

## Approaches to automating classification

We formulate the chemical classification task as assigning a chemical **structure** to a chemical **class**, based on a formal definition or model of that class. Here, a structure is denoted by a string of characters representing its molecular composition and connectivity. These structural strings must be formatted in one of the well defined chemical structural representation systems, such as SMILES (Simplified Molecular Input Line Entry System)[9,10] or InChI (International Chemical Identifier) [11]. An example of this classification task would be classifying the structure of **artemisinin** (CHEBI:223316) as a **terpenoid** (CHEBI:26873) due to its isoprene unit composition (**Fig. 1**) [12]. Most classification systems are polyhierarchical, meaning a structure can potentially be classified into multiple classes, and this is true of the classifications in ChEBI. Here we also assume the “true path rule” applies to chemical classification. This rule, originally from the Gene Ontology, was originally stated as “If the child term describes the gene product, then all its parent terms must also apply to that gene” [13]. Under this rule, when a structure **S** belongs to a class **C2**, and **C2** is an *is-a* child of **C1**, then **S** also belongs to class **C1**.

A number of approaches have been proposed or applied for chemical classification, including logical or ontological reasoning, cheminformatics approaches, and deep learning. Below, we describe these approaches and discuss their strengths and weaknesses.

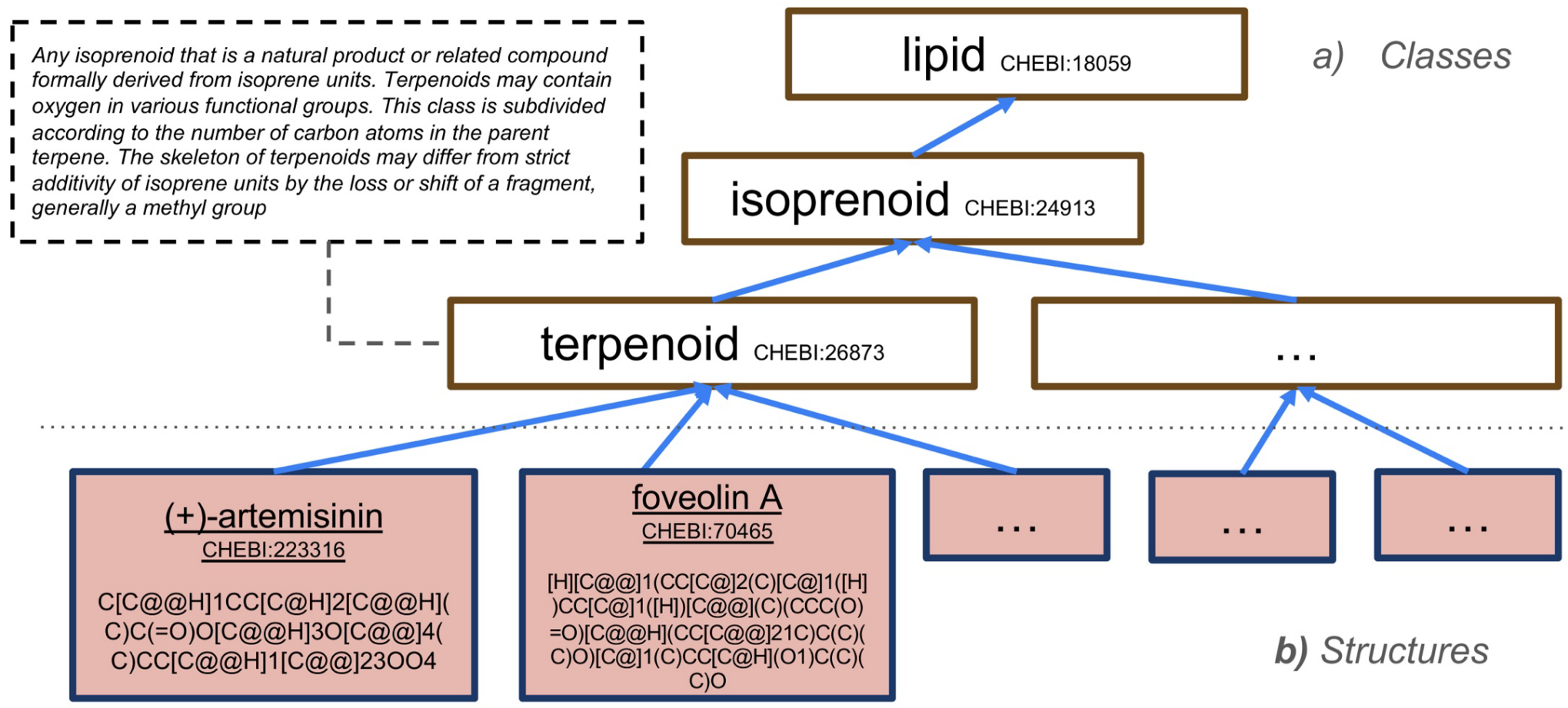

## Figure 1: Example of chemical structure classification

A subset of the ChEBI ontology, with terms organized in a hierarchy. **a)** Chemical classes serve as groupings for individual structures. These are arranged in a parent-child hierarchy, with more general classes at the top (e.g. every terpenoid is an isoprenoid, which is a lipid). Classes may be accompanied by natural language definitions that describe class membership conditions (the definition for *terpenoid* is shown). **b)** Beneath the classes, the leaf nodes typically correspond to discrete *structures* (e.g. specific stereoisomers of artemisinin (CHEBI:223316)), with ground (non-wildcard) SMILES strings.

### Description Logic and First Order Logic Reasoning Approaches

Description Logics (DLs) are a formalism for representing the criteria for class membership, allowing for the use of automated DL reasoners to automate classification tasks such as subsumption (subclass relationships between classes) or determining if an instance belongs to a class. For example, the class **sesquiterpenoid** can be logically defined as a **terpenoid** that has exactly 3 isoprene unit components; using OWL Manchester syntax[14] this would be written as as "**sesquiterpenoid** *EquivalentTo* **terpenoid** *and hasComponent exactly* **3 'isoprene unit'**".

Formally, DLs are a subset of full First Order Logic (FOL). The most widely used DL is the Web Ontology Language (OWL) [15], which also serves as the main exchange language for ontologies such as ChEBI. DL reasoning has been widely employed by a number of other ontologies such as the Gene Ontology[16], the Uberon anatomy[17], the Cell Ontology[18], and several phenotype and disease ontologies[19]. It is used as part of many ontology release pipelines to automatically classify portions of the ontology hierarchy[20].

In contrast to many ontologies, ChEBI does not employ DL reasoning to automate any aspect of structure classification. One reason for this is the lack of expressivity of OWL-DL, which has limited ability to specify higher-order relationships[21].

Formalisms such as Description Graphs have been proposed to overcome this[22,23], but these still lack the expressive capabilities necessary for full chemical classification. More recently, there has been work on using more expressive FOL approaches to define ChEBI classes and automate associated classification. An example of this is translating SMILES strings with wildcards and R-groups into FOL Rules, which was demonstrated to find misclassifications in ChEBI [24]. However, extending this mechanism to more general classes that lack R-groups would involve time-consuming authoring of FOL rules, requiring both logic and chemistry expertise. Additionally, these rules may be low-level and unintuitive for complex classes. Moreover, FOL model checking is known to be undecidable in general [25], meaning that not all automated reasoning problems in this formalism will finish in a finite amount of time and timeouts would have to be applied.

### SMARTS-based rule classification

SMARTS patterns[26] generalize motifs found in chemical structures with regular-expression-like features, and offer a more natural way to capture structural features compared to OWL-DL. These SMARTS expressions can be run against a SMILES string representing a specific chemical structure to determine if the structure is matched or is not matched by a particular pattern. One simple example of a SMARTS pattern is **CCC[N,O]**, which would match any SMILES representing a structure with three consecutive carbons followed by either a nitrogen or an oxygen [27]. The ClassyFire system[28] uses curated SMARTS patterns to define 4,825 hierarchically organized chemical categories. Because SMARTS strings alone are not expressive enough to represent most classes, the ClassyFire system leverages over 9,000 SMARTS patterns with boolean and programmatic combination to determine what category a chemical structure belongs to (although these SMARTS patterns and their combinations are not publicly available, so the precise system is unknown to us).

The ClassyFire API is leveraged by both ChEBI curators and PubChem to classify chemical structures. When used in ChEBI curation, a mapping and manual validation step is involved – ClassyFire uses its own ontology called ChemOnt, which makes its usage for classifying ChEBI challenging, since the fundamental classification strategies differ between the ontologies and as a result mappings are imprecise[29].

The ClassyFire/ChemOnt system is currently static and has not been updated since 2016: creating new classes in ChemOnt would require manual construction of new SMARTS string boolean expressions, and there is no process in place for doing this. Moreover, the existing SMARTS expressions are not openly available, hindering potential community contributions.

### Deep learning approaches

A number of approaches have been used that leverage deep learning neural network (NN) approaches, employing vector embeddings of SMILES strings[30]. The current state of the art for this approach is Chebifier[31]: this method learns latent chemical representations from training data, which can then be used to predict class membership for new structures. Whereas symbolic rule-based approaches such as OWL-DL and SMARTS rules are manually curated, NNs can be trained directly from existing data, and Chebifier achieves high accuracy (achieving a micro F1 score of 90% on classifying individual molecules to their ChEBI class). However, unlike symbolic approaches, the underlying latent array-based representations are difficult to interpret and explain in terms of human-understandable chemical features, although limited forms of interpretability can be assigned to such networks in some cases[32]. Additionally, Chebifier's macro F1 score is only 66%, largely due to imbalances in class sizes in ChEBI, resulting in much poorer predictive performance for classes that are not well represented in the training data.

### Direct classification using generative AI

One potential approach to automating classification is the use of generative AI, in particular natural language-based Large Language Models (LLMs)[33]. LLMs have been used successfully to perform many different kinds of tasks, such as summarization, information extraction, and question answering[34]. The performance of LLMs in different fields is often measured by benchmarks; however, we are not aware of comprehensive benchmarks for the chemical classification task. In chemistry, the comprehensive ChemEval suite[35] includes a large variety of different evaluation benchmarks, but includes no benchmarks for chemical classification.

We have previously evaluated the use of LLMs and Retrieval Augmented Generation (RAG) approaches to classify biomedical ontology terms as part of our DRAGON-AI evaluation[36]. We evaluated two approaches for the task of classification (1) direct classification of terms; (2) generation of OWL classification axioms, which can then be used by OWL-DL reasoners. We evaluated these approaches over ten ontologies (this evaluation did not include ChEBI). The first approach demonstrated moderate accuracy, but would be challenging to scale to large chemical databases, as each structure would require an expensive LLM inference step. The second approach requires LLMs only to build class-level logical models, with no runtime classification dependency on LLMs, but is limited by the expressivity of OWL-DL, as noted above.

### Opportunity: Generative AI for classification program synthesis

Another potential approach is to use LLMs to generate customized classification programs for each class in the ontology. Each program, implemented as code in an executable programming language such as Python, would test whether a given chemical structure (as defined by its SMILES string) fits into the particular class. An example program for classifying members of the class “alkane” is shown in **Fig. 2.** These programs would have the ability to call software libraries such as the Python RDKit library[37] or OpenBabel[38]. The RDKit includes functions for SMARTS-based classification, and is already used by major chemical entity databases such as ChEMBL to assist in curation by performing tasks such as standardizing structures and deriving parent structures[39].

Manually curating individual programs for the thousands of grouping classes in ChEBI would be an expensive and time-consuming process, requiring expertise in both chemistry and programming. But this could be rapidly accelerated via the use of LLMs, which have shown proficiency in program synthesis tasks[40]. This can also be seen as a natural extension of our approach to generate OWL-DL definitions using DRAGON-AI, replacing the limited expressivity of OWL-DL with more expressive Python programs.

```
from rdkit import Chem

def is_alkane(smiles: str) -> tuple[bool, str]:
    """
    Determines if a molecule is an alkane based on its SMILES string. An alkane is characterized by the
   formula CnH2n+2 and consists of saturated carbon atoms and hydrogen atoms, without any rings.
    """
    # Parse SMILES
    mol = Chem.MolFromSmiles(smiles)
    if mol is None:
        return False, "Invalid SMILES string"

    # Check for presence of only carbon and hydrogen atoms
    elements = {atom.GetAtomicNum() for atom in mol.GetAtoms()}
    if elements.difference({6, 1}):  # Atomic number 6 is C, 1 is H
        return False, "Contains atoms other than carbon and hydrogen"

    # Check if the molecule is saturated (only single bonds)
    for bond in mol.GetBonds():
        if bond.GetBondTypeAsDouble() != 1.0:
            return False, "Contains unsaturated bonds (double or triple bonds present)"

    # Check for acyclic structure (cannot have rings)
    if mol.GetRingInfo().NumRings() > 0:
        return False, "Contains rings, not acyclic"

    # Count carbons
    carbon_count = sum(1 for atom in mol.GetAtoms() if atom.GetAtomicNum() == 6)

    # Calculate the total hydrogen count, including implicit hydrogens
    hydrogen_count = sum(atom.GetTotalNumHs() for atom in mol.GetAtoms() if atom.GetAtomicNum() == 6)

    # Calculate expected hydrogen count based on alkanes' CnH2n+2 rule
    expected_hydrogen_count = 2 * carbon_count + 2

    if hydrogen_count != expected_hydrogen_count:
        return False, f"Formula C{carbon_count}H{hydrogen_count} does not match CnH2n+2 (expected
H{expected_hydrogen_count})"

    return True, "Molecule matches the definition of an alkane"
```

## Figure 2. An example of a Python program to classify alkanes

This manually created exemplar program is a single function that takes as input a SMILES string, and returns a tuple of a boolean (true/false, indicating membership in the class), and a text string with an explanation of the decision. This example program uses the RDKit library to analyze the structure, calculating the number of bonds and rings, and integrating these sub-calls in order to arrive at a final classification. If a structure is not a member of a class, a reason is given (e.g. "Contains atoms other than carbon and hydrogen"). Both the internal structure of the program, with clear boolean logic and structural groupings, and the classification outputs, aim to be transparent and explainable.

### Contribution

In this manuscript, we evaluate an approach for generating a suite of classifier programs called C3PO (ChEBI Chemical Classifier Programs Ontology). The resulting programmatic ontology can be used for deterministic classification of SMILES structures, without any runtime dependency on an LLM or machine learning library. Furthermore, the programs and their results are transparent and explainable, intended to engender trust and easy verification, as well as allowing for future evolution of the rules by both human experts and other machine agents.

# Methods

## Iterative learning chemical classifier programs

### Components

Our approach for synthesizing chemical classifier programs has the following components:

- **An instruction-tuned text-based large language model (LLM)**: Either general purpose or coder-style LLMs can be used here, as well as newer reasoner-style models. LLMs may be local, or external and called via API.
- **A Python Execution Environment (with RDKit pre-installed)**: This environment allows the execution of the generated Python programs (C3Ps); allowing these programs to leverage the RDKit library to perform chemical structure analysis and manipulation.
- A **benchmark dataset** of classified chemical structures, which is split for learning and testing. Each chemical class contains a textual definition, a set of positive instances of the class (i.e. a list of chemicals which are in the class) and a set of negative instances of that class (i.e., a list of chemicals which are not in the class).

### Learn-Execute-Iterate-Adapt Algorithm

We propose a method called LEIA (Learn-Execute-Iterate-Adapt) that generates a ChEBI Chemical Classifier Program Ontology (C3PO) using an iterative process inspired by genetic programming approaches (**Fig. 3**). For each class *c*, this will generate a program $p_c$ that takes as input a SMILES string *s*, and yields a tuple $<m,e>$ of a boolean *m* indicating whether *s* is classified as a member of *c*, plus a concise explanation *e* for the boolean value.

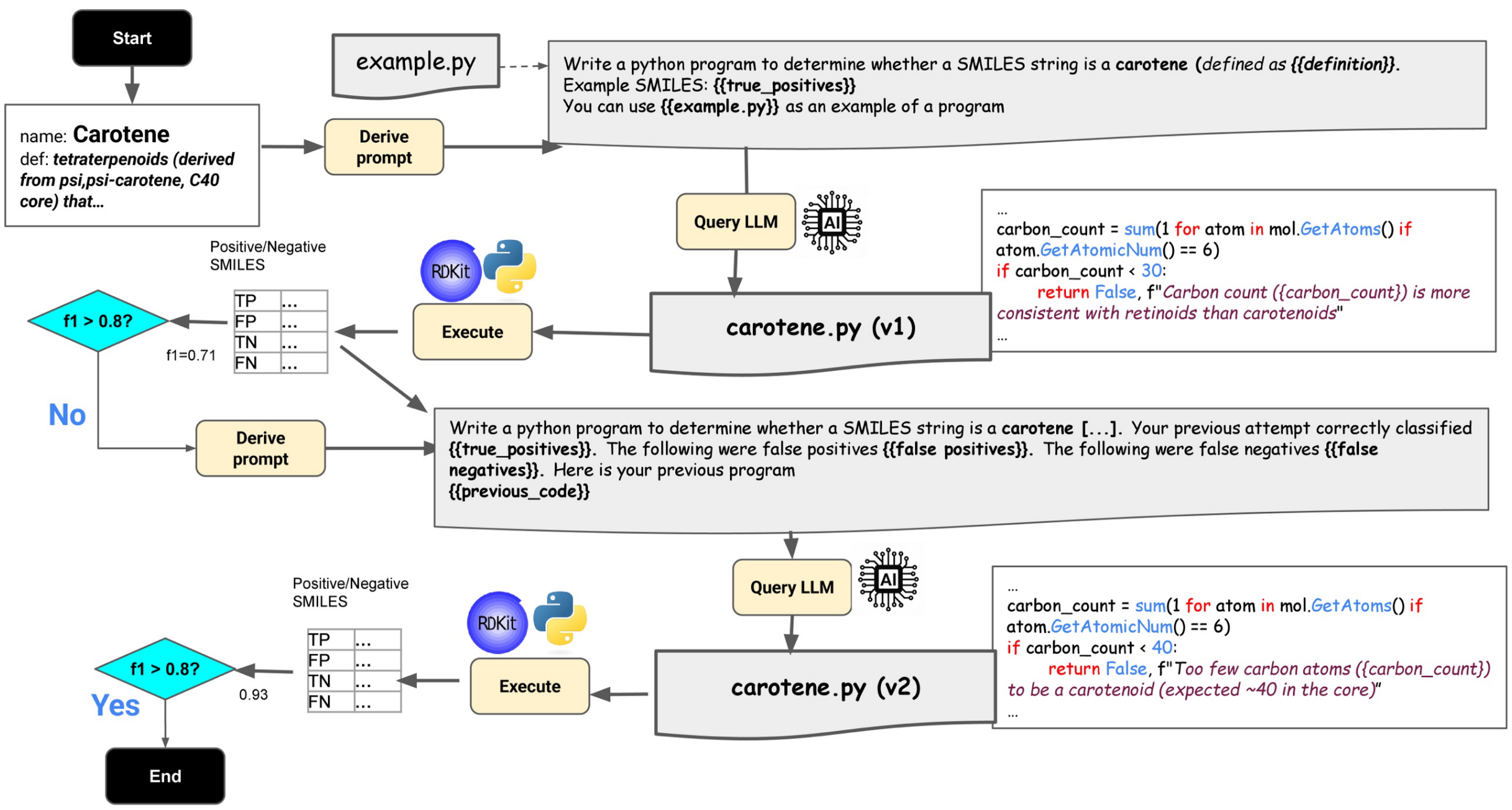

```
...
carbon_count = sum(1 for atom in mol.GetAtoms() if
atom.GetAtomicNum() == 6)
if carbon_count < 30:
    return False, f"Carbon count ({carbon_count}) is more
consistent with retinoids than carotenoids"
...
```

```
...
carbon_count = sum(1 for atom in mol.GetAtoms() if
atom.GetAtomicNum() == 6)
if carbon_count < 40:
    return False, f"Too few carbon atoms ({carbon_count})
to be a carotenoid (expected ~40 in the core)"
...
```

## **Figure 3**: **Learn-Execute-Iterate-Adapt (LEIA) procedure for generating classifier programs**

The starting point is a chemical class (in this case, carotene), a textual definition, and a set of positive and negative examples. LEIA prompts the LLM to generate the first version of the program; this is evaluated against the instance data. If this does not meet the F1 threshold, the LLM is prompted again, with feedback on the flaws of the program. The LLM continues to refine the program until either the threshold is reached, or the maximum number of iterations have elapsed. Note that this iterative process distinguishes this approach from a single-shot LLM approach.

### Inputs:

1. *Target Class Name/Identification:* The name of the ChEBI class for which a classifier program is to be generated.
2. *Target Class Definition*: The textual definition of the ChEBI class.
3. *Positive Examples*: A set of SMILES strings known to belong to the target ChEBI class.
4. *Negative Examples*: A set of SMILES strings known *not* to belong to the target ChEBI class.

### Parameters:

1. *F1 Threshold*: The minimum F1 score [41,42] required for the generated program to be considered successful. The default threshold can be modified, but for our evaluation we chose 0.8, based on initial empirical runs.
2. *Maximum Attempts*: The maximum number of iterations allowed for the refinement process, here set to 4.

3. *Model*: LLM model to be used, plus any parameters for that model.
4. *Prompt variants*: For example, whether the chemical class definition is to be included or excluded.

**Output:**

1. *C3P*: A Python program containing a function named `is_<class_name>`. This function takes a SMILES string as input and returns a two-argument tuple:
    - *m* - A boolean value indicating membership, whether the SMILES string belongs to the target ChEBI class.
    - *e* - A concise plain text explanation for the classification decision.

**Steps:**

1. **Prompt Generation:** Create a prompt to learn the program that includes:
    - Instructions for the LLM to generate a Python function that classifies SMILES strings into the target class, adhering to a specified format. The prompt requests the LLM to use Chain-of-Thought reasoning[43], and also to provide clear documentation in the program.
    - The target class definition.
    - A subset of the positive examples (both chemical name and SMILES).
    - An exemplar program; a manually constructed Python function that uses RDKit functions to perform classification for a single class.
    - Any additional text from subsequent iterations.
2. **LLM Prompt:** Submit the generated prompt to the instruction-tuned LLM. The LLM will generate a ChEBI chemical classification program (C3P) in Python based on the prompt. The response is parsed to separate out chain-of thought reasoning output from the Python code.
3. **Execution and Scoring**: Execute the generated C3P on the full set of positive and negative training examples using the Python execution environment with RDKit pre-installed. Evaluate its performance by calculating precision, recall, accuracy, and the F1 score. If there is a compilation or runtime error, capture the error and treat that F1 score as zero.
4. **Iteration:** If the F1 score of the C3P is below the F1 threshold, and the maximum number of attempts has not been reached, repeat steps 1-3 with an adjusted prompt, requesting the LLM to adapt the program. The adjusted prompt includes (a) the program itself (b) misclassified examples (the SMILES and chemical name as well as the explanation provided) (c) any runtime errors encountered.
5. **Termination**: The process terminates when one of the following conditions is met:

    - The generated C3P achieves an F1 score equal to or greater than the F1 threshold.
    - The maximum number of attempts is reached.

The resulting C3P represents a program that attempts to classify SMILES strings into the target ChEBI class with the desired level of accuracy. The programs may be filtered based on whether the desired threshold is reached.The procedure can be executed over a desired set of chemical classes for which instance data is available, creating a program for each class. The resulting set of programs is called a C3P *Ontology*, because it encapsulates formal representations of chemical classes that can be used for inference. Like a formal ontology specified in OWL-DL, it can be used for run-time classification of structures into classes.

### Inference over C3PO to classify de-novo structures

The resulting C3PO suite of programs can be used for runtime classification of de-novo SMILES strings. Given a SMILES structure *s*, for each class *c*, run the corresponding program $p_c$ . This yields a boolean true/false result, plus explanation. A confidence can be attached based on the performance characteristics of p(c) calculating from the above training steps as follows:

- If the classification is **True**, the confidence is the precision of $p_c$ (i.e *tp/(tp+fp)*)
- If the classification if **False**, the confidence is the negative predictive value of $p_c$ (i.e. *tn/(tn+fn)*)

Note that this confidence score may vary in some cases depending on the training data split and whether the training data accurately reflects the distribution from which the de novo molecule is sampled.

To allow for easy utilization of C3PO, we also provide a command line script that can be easily installed from PyPI that will provide classification of SMILES strings passed as arguments.

### Tracking evolution of programs using version control systems

For each experiment, we created a version controlled folder using the git version tracking system, and we create a commit for every iteration through the learning procedure. The model's chain of thought is included as comments attached to the commit, and for all iterations subsequent to the initial iteration, the feedback we provide the model is also included as a git commit. The history is published on GitHub, making the evolution of each program transparent and easily explored using normal GitHub navigation features.

## Generation of C3PO Benchmark

The benchmark dataset used to evaluate the performance of C3PO is derived from the ChEBI ontology. ChEBI, like most ontologies, consists entirely of entries that are modeled as *classes* in the OWL-DL representation. Here we use an alternative system, the CHEMROF framework [44], based on LinkML [45], which allows a candidate partitioning of ChEBI into *grouping classes* (or simply "classes", in this manuscript), and *chemical entities* (the things to be classified, also called *structures* here).

We use the following approach to partition ChEBI entries into *structures* and *classes*:

- If an entry has a SMILES string, it is included as a *structure*. We eliminate SMILES strings that contain wildcards ("*"), as well as CHEBI entries with subclasses, focusing only on structures with non-wildcard containing SMILES strings (typically corresponding to leaf nodes).
- Otherwise, if the entry is a direct or indirect is-a parent (superclass) of at least one structure, it is included as a *class*. All structures grouped in this way are treated as positive examples, and all other structures count as negative examples.
- Other entries are discarded.

This procedure implicitly filters out ChEBI roles, which provide an alternative non-structural means of classifying structures[46] (because ChEBI roles are never superclasses of structures). This procedure also eliminates all obsolete entries, as these also are never superclasses. A subset of the classes have SMILES strings (either using wildcards or R groups), but most do not. It should be noted that many of the SMILES strings for classes in ChEBI are not unique and are only informative rather than definitional. For example, the ChEBI class for *ultra-long chain fatty acids* has a structure that is a generic carboxyl functional group (-COOH) attached to an unspecified R group. This represents necessary but not sufficient criteria. Because there are many such cases we ignore any structural formulae attached to classes.

The resulting dataset is available on Hugging Face[47].

For evaluation purposes, we filtered down the set of ChEBI classes to those meeting the following criteria:

- Minimum of 25 members (to have a sufficient number for testing)
- Maximum of 5000 members (to eliminate very high level groupings)
- Has a textual definition in ChEBI (to help eliminate ambiguity)

ChEBI also assigns curation ratings (between 1 and 3 stars, with the majority being 2 or 3 stars). In order to maximize the size of the training and test set, we include all structures regardless of rating.

We executed this procedure on ChEBI v237 to create the core C3PO benchmark, consisting of 1,364 classes and 177,875 structures. The 177,875 structures are split into 80% for learning and 20% that is held back for validation.

The resulting dataset includes many ChEBI classes that are rarely used in the biochemical literature, so we then created a biologist "slim" of C3POv237, intended to capture the most biologically relevant subset. To do this, we used mappings[48] as a proxy for biological relevance, on the assumption that classes that are mapped to metabolomics identifiers or identifiers in pathway databases such as KEGG[49] are more likely to be relevant. ChEBI also includes mappings to PubMed and Wikipedia, and we assume these correspond more to relevant classes.

This resulted in 346 classes in C3PO-slim. We then define a further subset, consisting of classes which Chebifier has been trained on (C3PO-slim-chebifier), taking the number down to 242.

Note that due to the minimal member criteria, most members of C3PO are polyatomic molecules of varying degrees of complexity. There are also a small number of atoms such as “polonium atom” which end up being included due to the fact that a large number of distinct isotope entries are enumerated in ChEBI.

## Evaluation against the ChEBI benchmark

We evaluated the following models:

- gpt-4o, a standard OpenAI model
- gpt-o1, an OpenAI model aimed at reasoning tasks
- gpt-o3-mini, a more cost-effective reasoning model
- gemini-2.0-flash-exp
- deepseek-r1, an alternative reasoning model from the DeepSeek group. We used the full 671b parameter model, available via the TogetherAI API
- claude-sonnet-3.5, from Anthropic

In all the above cases, we set the value for min F1 threshold to be 0.8, and max iterations to be 4. We tested a variant of these hyperparameter settings on gpt-4o and o3-mini where the F1 threshold was 0.9, and the maximum number of iterations was 6. We also tested a variant in which the prompt included instructions to avoid assuming that the provided positive and negative examples were all correct, which we call “use-the-force” mode, intended to allow the LLM to override what it believes to be incorrectly classified examples in the training set.

For each experiment, we learned a C3PO suite for all 346 classes in C3PO-Slim, holding back all validation structures. We also created an *ensemble* program ontology by taking the best performing model (on training data) for each chemical class. This gave a total of 10 different experiment runs, and a total of 9 different candidate C3PO suites (see **table 1**).

We then evaluated each C3PO suite over all held-back structures in the validation subset . For further evaluation of performance, we also generated Chebifier classifications for all these structures (described in the Introduction section). Chebifier was called via its API  (executed on Jan 7, 2025)

As a baseline we also included a naive single SMARTS-based classification approach (“smartifier”), making use of the generalized SMILES strings present for a subset of the C3PO benchmark classes in ChEBI.  These are typically either wildcard-containing SMILES strings, or SMILES strings containing R groups, and can be treated as SMARTS patterns in RDKit. For every ChEBI class with a SMILES/SMARTS string, we created a naive classifier in which we use RDKit to check membership of the SMILES of the structure against the ChEBI SMARTS string.

For the main evaluation, we only used classes in C3PO-Slim that are also learned by Chebifier, giving 242 classes. This further slimming is necessary to give an accurate comparison. We also performed comparisons that exclude Chebifier over all 346 classes, available as supplementary data.

For each metric (F1, precision, recall, accuracy), we calculate both macro and micro versions. For macro, we calculate each metric on a per-class metric, and then calculate the mean for all evaluated classes. For micro, we sum all individual outcomes and calculate metrics from this.

| Experiment | Model | Use the force | Max iterations | Min F1 |
|---|---|---|---|---|
| claude-sonnet | claude-sonnet | no | 4 | 0.8 |
| gpt-4o | gpt-4o | no | 4 | 0.8 |
| o1 | o1 | no | 4 | 0.8 |
| o3-mini | o3-mini | no | 4 | 0.8 |
| deepseek-r1 | deepseek-r1 | no | 4 | 0.8 |
| gemini-2.0-flash-exp | gemini-2.0-flash-exp | no | 4 | 0.8 |
| claude-sonnet-force | claude-sonnet | yes | 4 | 0.8 |
| gpt-4o-iter6 | gpt-4o | no | 6 | 0.9 |
| o3-mini-iter6 | o3-mini | no | 6 | 0.9 |
| *ensemble* | *all* | mix | 4-6 | 0.8-0.9 |

**Table 1. All experiments, with the models and hyperparameters used.**

## Molecular descriptor analysis

In order to explore the relationship of molecular properties to performance of the system, we defined a set of 37 derived molecular descriptors, and implemented these in RDKit. Descriptors were categorized into: (1) basic molecular properties (molecular weight, ring counts), (2) drug-likeness metrics (hydrogen bond donors/acceptors, topological polar surface area), (3) functional group counts using SMARTS pattern matching, (4) structural complexity measures (chiral centers, bridgehead atoms), and (5) class-specific markers for lipids, carbohydrates, and steroids.

For lipid detection, we implemented specialized SMARTS patterns to identify glycerol backbones ([OX2][CX4H2][CX4H]([OX2])[CX4H2][OX2]) and acyl-glycerol linkages. Steroid detection used a four-ring fused system pattern, while carbohydrate markers included pyranose/furanose rings and glycosidic bonds. Custom algorithms calculated the longest aliphatic chain length using breadth-first search on non-aromatic carbons and counted fused ring pairs through ring intersection analysis.

These descriptors are defined in **Supplementary table S1**.

In order to evaluate the relationship between each descriptor and program learnability, we characterized each learned class by the median descriptor value across all member structures. To identify descriptors that distinguish performance between methods, we conducted Mann-Whitney U tests comparing classes where C3P (using the ensemble programs) outperformed Chebifier (F1 difference > 0.05) versus classes where Chebifier excelled. Multiple hypothesis correction was applied using the Benjamini-Hochberg false discovery rate (FDR) procedure with $\alpha = 0.05$.

Effect sizes were quantified using Cliff's delta ($\delta$), interpreted as: $|\delta| < 0.147$ (negligible), $< 0.33$ (small), $< 0.474$ (medium), $\geq 0.474$ (large). Negative $\delta$ values indicated higher descriptor values in Chebifier-favored classes, while positive values indicated C3P-favored classes.

For individual method analysis, we categorized classes into performance groups based on F1 scores: poor (<0.5), moderate (0.5-0.7), and good (>0.7). Mann-Whitney U tests with FDR correction compared descriptors between good and poor performing classes to identify molecular features associated with method success or failure.

# Results

## Programs can be learned for a subset of chemical classes

We used the LEIA approach to a synthesize classifier program ontology (C3POs) for all 346 classes in the benchmark, using 9 different hyperparameter settings (see methods), against a subset of the C3PO benchmark with 20% of structures held out for validation. We also generated an ensemble program suite, which included the best performing program for each class on training data (C3PO-ensemble).

For each class in the benchmark, we define the *learnability* as the F1 score for that class evaluated against the training set. The classes in the benchmark showed a wide variety of learnability scores, with a mean of 0.465, and 28% of classes having programs that scored above 0.7. The top scoring and bottom scoring classes are shown for the ensemble model in **Fig. 4** (for performance across models, see **Fig. S1**). The top ten learnable classes include both trivial atom-level classifications (chalcogen, polonium atom), as well as many glycerol backbone based classes (cardiolipins, triglycerides, triradylglycerol). The least learnable classes included more complex categories, including terpenoids and bioactive molecules such as ‘dihydroagarofuran sesquiterpenoid’, as well as the general grouping class of ‘bioconjugate’).

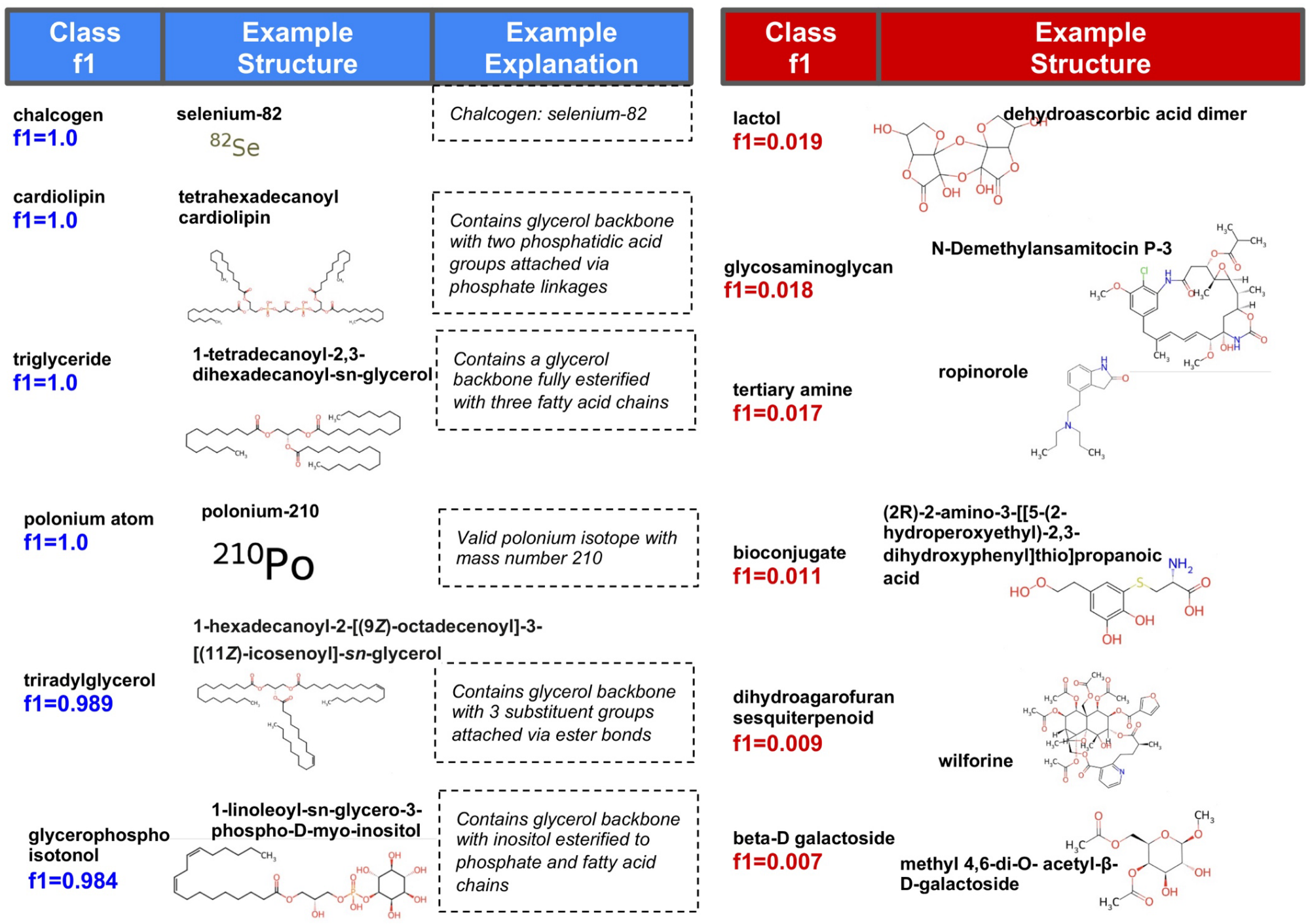

## Figure 4. Highest and lowest scoring classes from the learning phase

The left panel shows top learned chemical classes by F1 during training (ensemble C3PO). This includes a mixture of trivial classes (e.g. atoms or atom groupings) plus more complex structures often characterized by glycerol backbones (in the case of the most learnable classes, there is some overlap). We show some example structures positively classified using these programs, together with the verbatim ‘explanation’ output of the program. The right panel shows classes that had poorest learnability, i.e. those with the lowest F1 scores during training. We show some examples of these classes: these are typically more complex biological molecules, natural products, and highly broad non-structural categories.

In order to systematically identify features that correlate with learnability, we defined 37 derived molecular descriptors (see methods), and looked for enrichment between these features and the learnability score. Learnability was positively correlated with features such as high glycerol backbone count, and long aliphatic chains. Learnability was negatively correlated with the number of rings in the structure, indicating either a systematic inability for LLMs to derive reliable programs involving reasoning over ring systems, or potentially systematic errors in the benchmarks for these classes. These findings are summarized in **Fig. 5**.

When evaluated on the held-out test data, the learnability of the model was a strong predictor of performance (r=0.91; see **Fig. S2**).

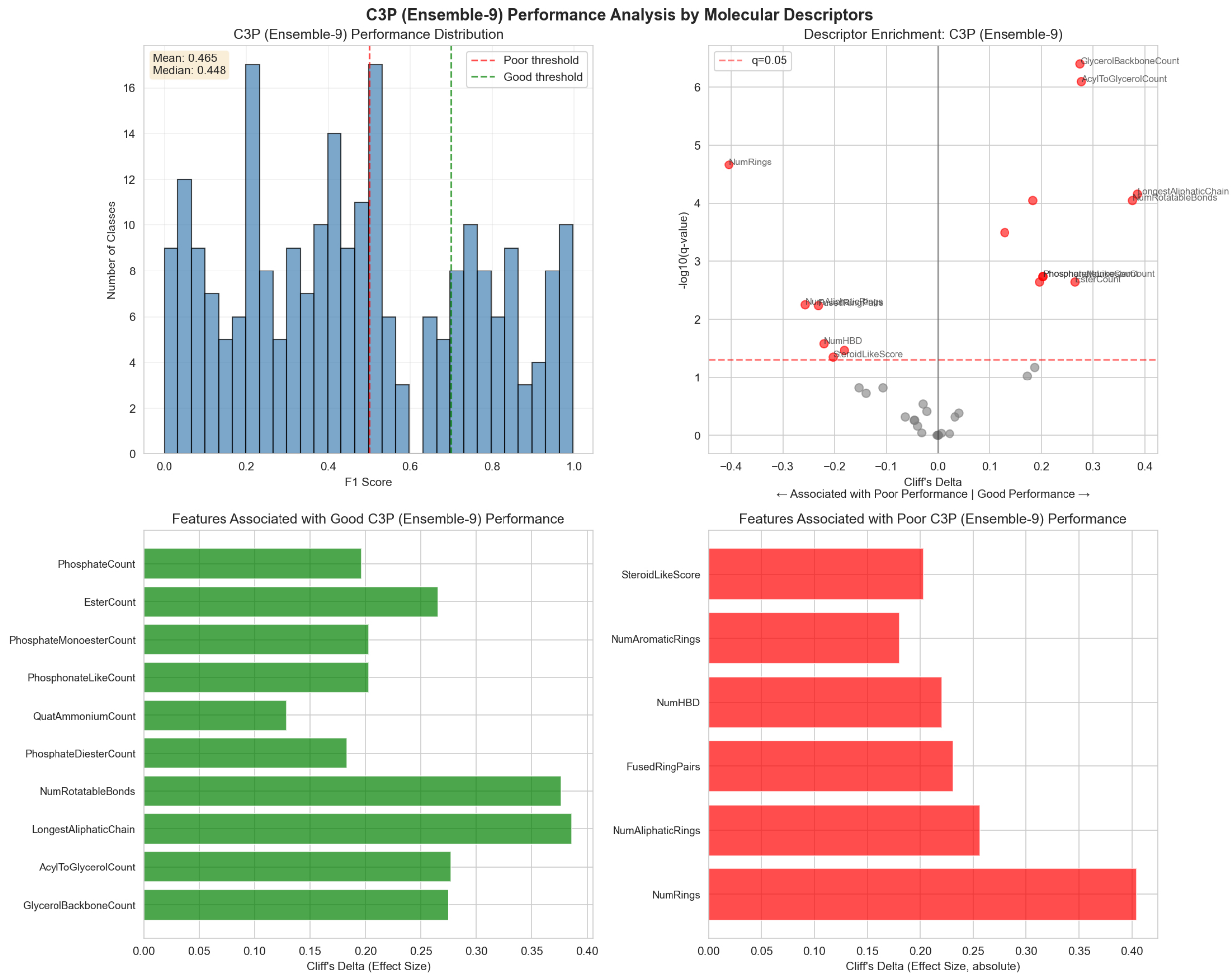

## Figure 5: Features that affect learnability

Analysis of which molecular descriptors have positive or negative effects on learnability. Top left: distribution of per-class F1 scores from learning phase (learnability). Bottom left and right: molecular descriptors that positively correlate (left) or negatively correlate (right) with learnability, scaled by effect. Top right: molecular features, effect size (Cliff's delta) vs significance (q-value). Overall glycerol backbones associated with good learnability (right side of figure), number of rings with poor learnability (left side).

An example of a learnable class is *glycerophosphocholine*, a class of chemicals with important roles in human health. The final learned program for this is shown in **Fig. 6**. This is a relatively simple program, consisting of a logical disjunction of three SMARTS strings. This program achieved F1 of 0.974 on the training data, and 0.970 on the test dataset.

```python
from rdkit import Chem

def is_glycerophosphocholine(smiles: str):

    mol = Chem.MolFromSmiles(smiles)
    if mol is None:
        return False, "Invalid SMILES string"

    # List of SMARTS patterns to match the glycerophosphocholine connectivity.
    # Pattern 1: fully free glycerol backbone.
    pattern_free = "OCC(O)CO[P](=O)(O)OCC[N+](C)(C)C"
    # Pattern 2: lysophosphatidylcholine with one acyl chain at position sn-1.
    pattern_lyso_sn1 = "OC(=O)OCC(O)CO[P](=O)(O)OCC[N+](C)(C)C"
    # Pattern 3: lysophosphatidylcholine with one acyl chain at position sn-2.
    # In the SMARTS below, the '*' after OC(=O) allows for any carbon chain.
    pattern_lyso_sn2 = "OCC(OC(=O)*)CO[P](=O)(O)OCC[N+](C)(C)C"

    patterns = [
        (pattern_free, "Found free glycerol backbone connectivity"),
        (pattern_lyso_sn1, "Found lysophosphatidylcholine connectivity (acyl at sn-1)"),
        (pattern_lyso_sn2, "Found lysophosphatidylcholine connectivity (acyl at sn-2)")
    ]

    for patt_str, explanation in patterns:
        patt = Chem.MolFromSmarts(patt_str)
        if patt is None:
            continue
        if mol.HasSubstructMatch(patt):
            return True, explanation

    return False, "Glycerophosphocholine connectivity not found; either phosphocholine
      headgroup or proper glycerol linkage is missing or has too many acyl substitutions."
```

**Figure 6: learned program for glycerophosphocholine (ensemble model)**

The program has a relatively simple structure, combining three different SMARTS patterns in a disjunctive (OR) pattern, with the name of the SMARTS pattern included in the classification. This program scored well on the training set (F1 of 0.917).

We also investigated the relative performance of each individual model contributing to the ensemble program set, as well as different hyperparameter settings. **Fig. 7** and **Table 2** shows the distribution of F1 scores on held-back test data across all 9 experiments, plus the baseline SMARTS approach ("SMARTifier"). In all the base experiments, OpenAI models with an emphasis on reasoning outperformed other models, with o3-mini performing the best (F1) of all the pure LLM models. The C3PO ensemble model over all 9 experiments is overall best, with a significant boost over the top base experiment.

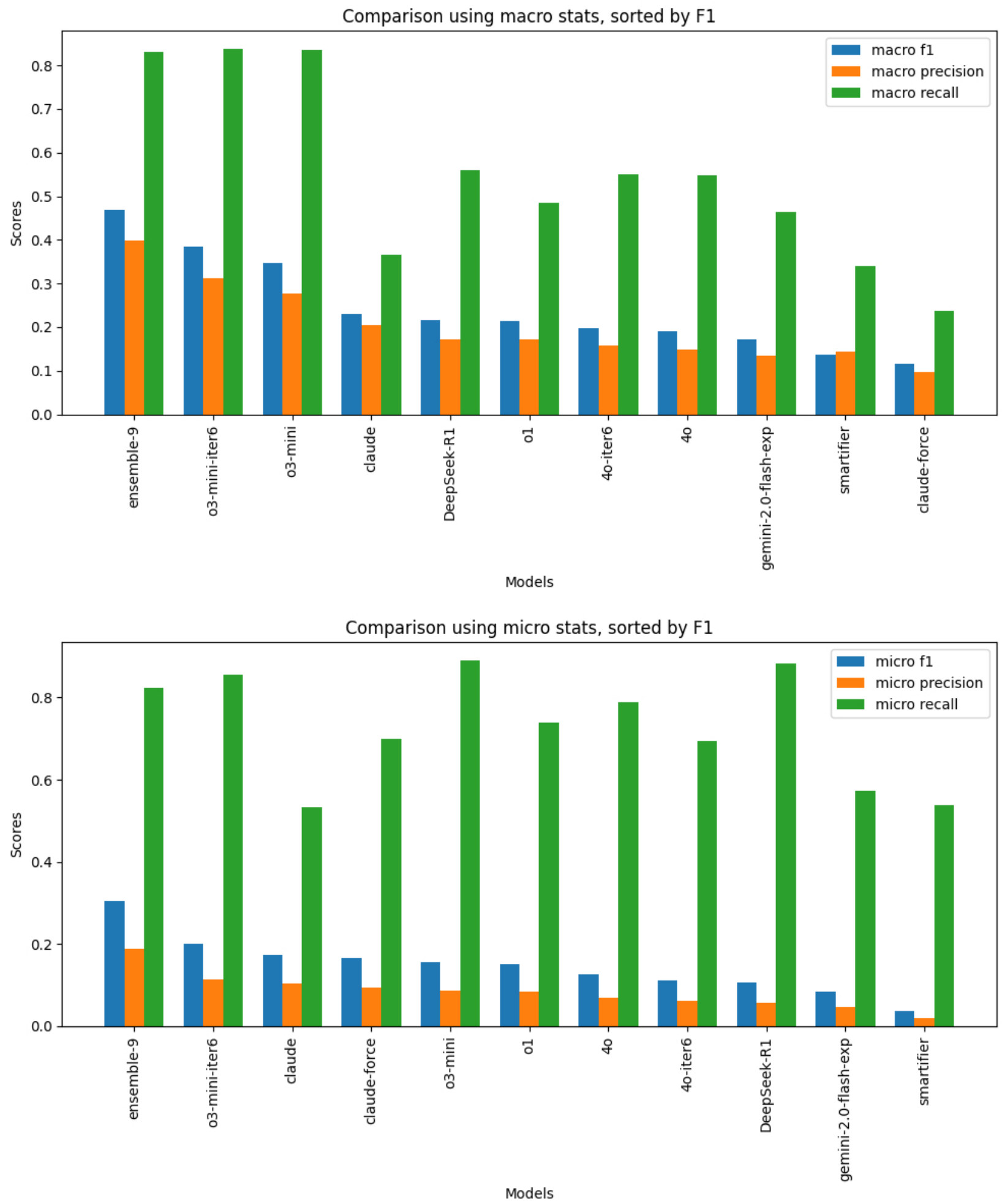

## Figure: 7. Model Comparison

Macro and micro outcome statistics for all C3POs from all 9 experiments, plus the ensemble C3PO. We also include the naive Single-SMARTS-based classifier (labeled “smartifier”). Iter6 means that LEIA was permitted up to 6 iterations, and an F1 threshold of 0.9. ‘Force’ indicates that the model was allowed to prioritize its own judgment over positive and negative examples. After the ensemble programs, the best scoring programs are those from the o3-mini model, with up to 6 learning iterations.

| **Macro-averaged Metrics** | | | |
|---|---|---|---|
| **Method** | **F1** | **Precision** | **Recall** |
| **c3po (ensemble)** | 0.469 | 0.399 | 0.831 |
| o3-mini (6 iterations) | **0.384** | **0.311** | **0.837** |
| o3-mini | 0.346 | 0.277 | 0.835 |
| claude-3-sonnet | 0.231 | 0.205 | 0.366 |
| deepSeek-R1 | 0.216 | 0.173 | 0.559 |
| o1 | 0.214 | 0.173 | 0.486 |
| gpt-4o (6 iterations) | 0.198 | 0.159 | 0.551 |
| gpt-4o | 0.191 | 0.149 | 0.548 |
| gemini-2.0-flash-exp | 0.172 | 0.135 | 0.463 |
| smartifier | 0.137 | 0.145 | 0.34 |
| claude-3-sonnet (with use-the-force option) | 0.117 | 0.096 | 0.237 |

## Micro-averaged Metrics

| Method | F1 | Precision | Recall |
|---|---|---|---|
| **c3po (ensemble)** | 0.304 | 0.187 | 0.823 |
| o3-mini (6 iterations) | **0.201** | **0.114** | 0.856 |
| claude-3-sonnet | 0.174 | 0.104 | 0.532 |
| claude-3-sonnet (with use-the-force option) | 0.166 | 0.094 | 0.699 |
| o3-mini | 0.156 | 0.085 | **0.89** |
| o1 | 0.151 | 0.084 | 0.738 |
| gpt-4o | 0.125 | 0.068 | 0.788 |
| gpt-4o (6 iterations) | 0.112 | 0.061 | 0.695 |
| DeepSeek-R1 | 0.105 | 0.056 | 0.883 |
| gemini-2.0-flash-exp | 0.084 | 0.045 | 0.573 |
| smartifier | 0.037 | 0.019 | 0.537 |

**Table 2**: Outcome statistics for all experiments, with best scores for each outcome type shown in bold (excluding the ensemble model, which naturally has the highest F1, since it is constructed from programs that have the highest F1 for each chemical class).

Increasing the number of iterations gave a slight boost to performance for gpt-4o, as expected (indicated as 4o-iter6 in **Fig. 7**). We also investigated whether providing the model with ChEBI text definitions of the chemical classes helped or hindered, but this had no effect. For one experiment, we gave Claude instructions to trust its own judgment if some members of the training set did not classify as expected (the so-called “use the force” option, indicated as ‘claude-force’ in **Fig. 7**). We hypothesized that errors in the training set might lead the model to overfit or otherwise create suboptimal programs. However, using this setting had a largely detrimental effect.

## Program learning steps are traceable and interpretable

With traditional deep learning approaches, during the learning phase, weights in the network are updated through backpropagation. Neither these weights nor their updates can be easily interpreted by domain experts. In contrast, with the program learning approach, both programs, and the reasons for each update are interpretable.

To illustrate this, we created a GitHub repository containing the full iteration cycle for each learned program for each method. This allows us to explore how LEIA converged (or failed to converge) on a solution. An example of this is the evolution of glycerophosphocholine, which can be explored at https://github.com/chemkg/c3p/commits/main/c3p/programs/glycerophosphocholine.py and which is shown partially in **Fig. S3**. The initial attempt scored 0.89, and through subsequent iterations the score gradually increases, after a temporary setback on the penultimate attempt.

We also carried out an analysis of the properties of programs produced by different models (see **Fig. S4**), showing that more powerful models typically produce lengthier and more complex programs.

We also carried out an analysis of potential confabulations. LLMs are known to be prone to make up non-existent content[50], sometimes this is referred to as “hallucinations”. In code generation, this can sometimes manifest as inventing code libraries or functions in code libraries that do not exist[51]. We examined generated code for such non-existent functions, and we found an absence of such errors on the code generation task, with one exception. During the experiments with the Claude model, it frequently made up a function in the RDKit library called “rdDecomposition”. As far as we can tell, there is no such function in the RDKit library, nor has there even been such a function in the RDKit GitHub repository, or in released code.

## Learned programs are complementary to chemical structure based deep learning models

We evaluated against all held-out examples, and also evaluated against the state of the art deep learning method, Chebifier. For comparison we used the 243 classes common to Chebifier and C3PO-Slim. We calculated micro and macro outcome statistics for 3 metrics (F1, precision, recall). The comparison of the ensemble model against Chebifier is shown in **Fig. 8** and in **Table 3**, with full results in **Table S2**.

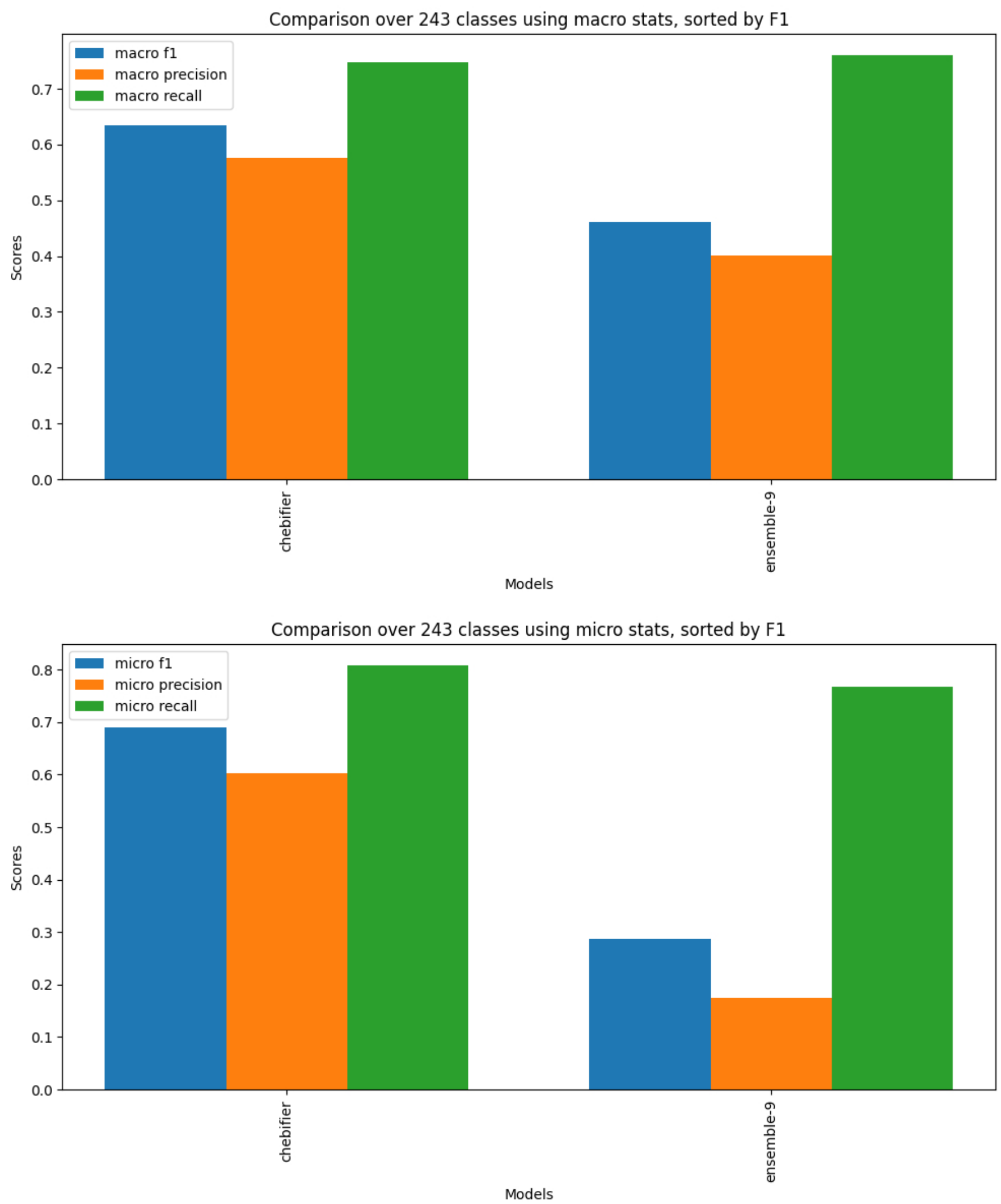

### Figure: 8. Comparison with Chebifier

Macro and micro outcome statistics comparing the ensemble C3PO against the state of the art deep learning classifier, Chebifier. When all classes in the benchmark are considered, Chebifier outperforms the C3PO approach on F1 and precision. The difference in performance with micro statistics is more pronounced, likely due to class imbalance.

| Macro-averaged Metrics |
| --- |

| Method | F1 | Precision | Recall |
|---|---|---|---|
| chebifier | **0.634** | **0.576** | 0.748 |
| c3po (ensemble) | 0.461 | 0.402 | **0.76** |

**Micro-averaged Metrics**

| Method | F1 | Precision | Recall |
|---|---|---|---|
| chebifier | **0.69** | **0.603** | **0.808** |
| c3po (ensemble) | 0.286 | 0.175 | 0.767 |

**Table 3**: Macro and micro statistics comparing the best learned programs (C3PO ensemble) against chebifier.

These results demonstrate that the deep learning approach retains a strong edge over the program learning approach when all classes are considered. This edge is particularly substantial when F1 is calculated as a micro statistic. This discrepancy between micro and macro statistics can be explained by the fact that when a class exhibits poor learnability, it consistently performs badly for all structures, and many of the classes that exhibit poor learnability have large numbers of structures. Using macro statistics evens out this imbalance.

To better understand the complementariness of the two approaches, we created a scatterplot of class-based F1 scores of both methods, shown in **Fig. 9**. There is a moderate positive correlation between the methods' respective scores, but there are multiple classes that are not near the diagonal on either side, indicating potentially divergent strengths of each approach. Note also that if we were to apply a learnability filter, the relative performance gap drops.

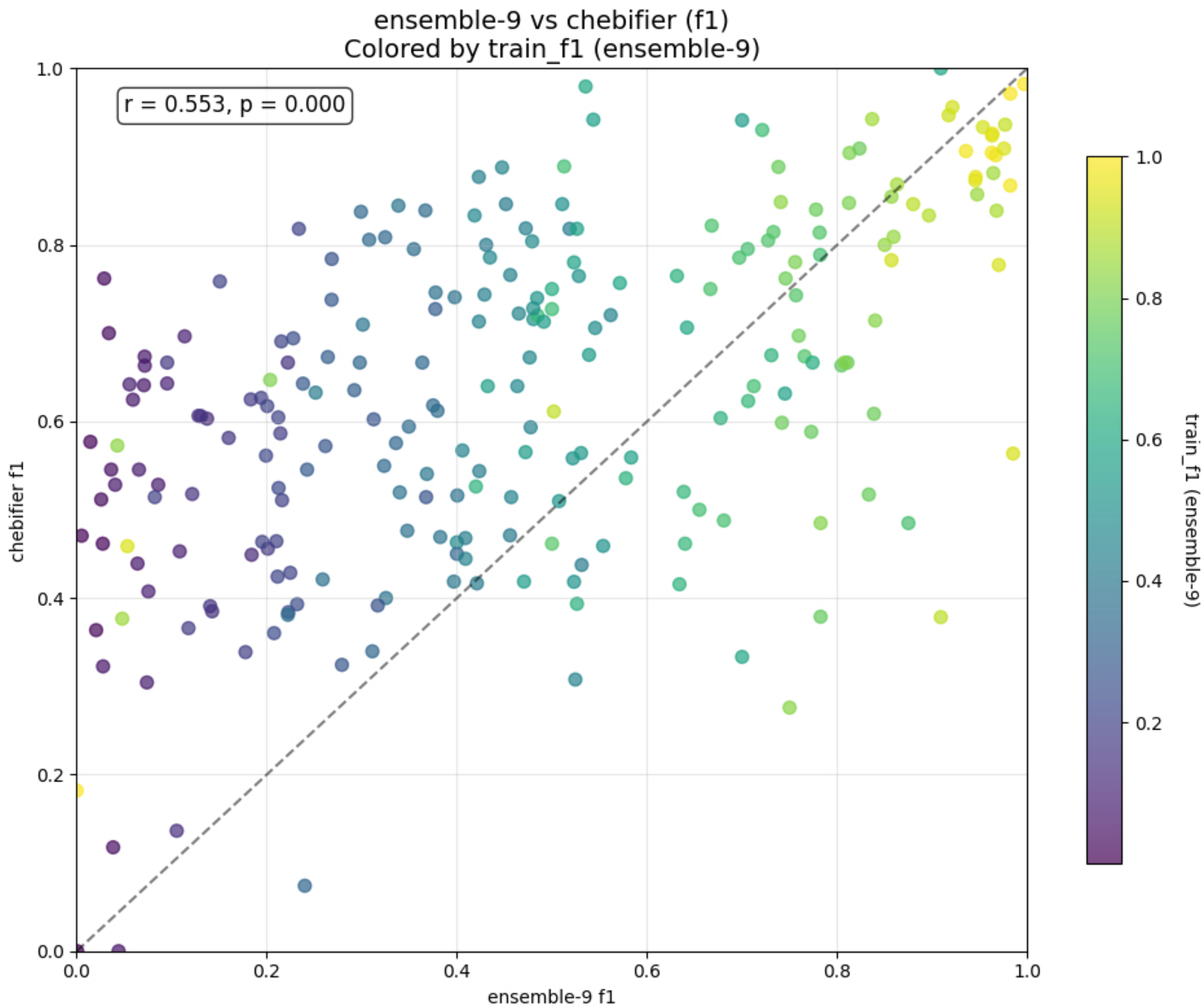

**Figure: 9. Per-class comparison**

Each circle represents a class, color coded by its learnability. Position of the class on x and y axes represents test-time F1 score on C3PO (ensemble) and Chebifier and respectively. F1 scores are moderately correlated, but the presence of outliers indicates potential strengths and weaknesses of each approach.

We investigated which properties contributed to the relative differences between the two approaches, using the same set of molecular descriptors defined in **supplementary table S1**. The enrichment analysis is shown in **Fig. 10**. This shows that the deep learning approach is not hampered by the number of rings in the same way that the program learning approach is. This also shows that the program learning approach has strengths when structures have multiple glycerol backbones. Other features like esther count are significant but show low effect. This feature based explanation can help suggest future directions to improve both approaches.

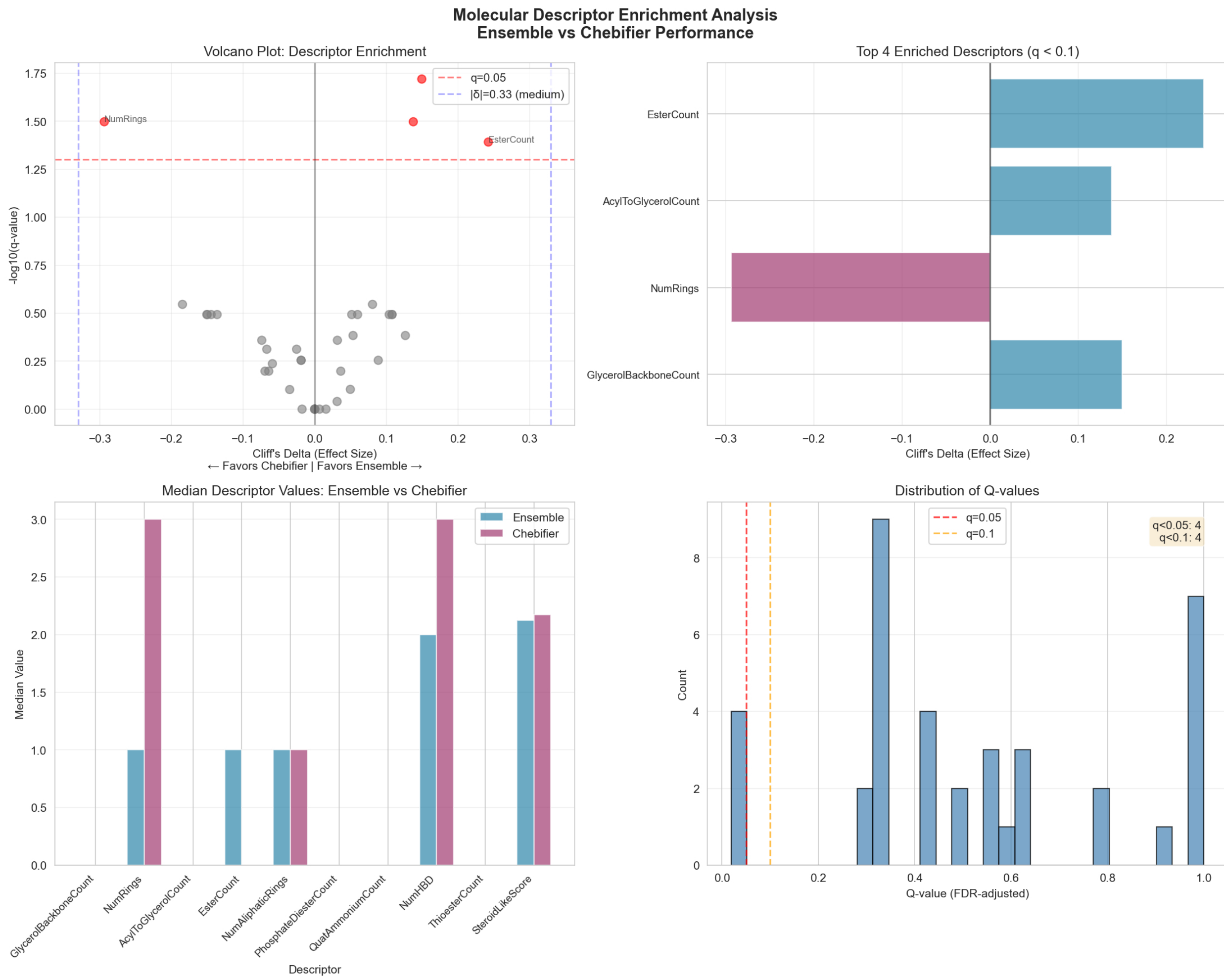

**Figure: 10. Analysis of molecular features driving difference between approaches**
Top left: volcano plot showing the effect and significance of each descriptor. Top right: top 4 descriptors, one of which (number of rings) negatively affects C3PO performance relative to Chebifier, the others exhibit a positive effect. Bottom left: Molecular properties where each method wins. Bottom right: distribution of Q values.

We manually examined a subset of structures in classes for which Chebifier and C3PO had divergent scores (**Fig. 11**.). An example is "1-monoglyceride", a class for which C3PO typically performs better than Chebifier. An example structure under this class is CHEBI:168508 with the CHEBI name "MG(18:3(6Z,9Z,12Z)/0:0/0:0)" and the IUPAC name "[(2S)-2,3-dihydroxypropyl] (6Z,9Z,12Z)-octadeca-6,9,12-trienoate". C3PO correctly classifies this, and gives the explanation "*Contains glycerol backbone with one ester bond at position 1 and two free hydroxyl groups*", which the curator rated as helpful.

In cases where C3PO gets the classification wrong, the explanation can pinpoint the source of the errors. For example, C3PO rejected "(2R,3S)-2-hydroxy-3-methylpentanoic acid" from the classification "2-hydroxy fatty acid", with the explanation "*Saturated fatty acid chain too short (5 carbons; require at least 8)*". We found this reason to be unusual, and when we examined it in more detail, we discovered an example in the training set that led the program learning to go awry; L-Isoleucic acid, a ChEBI 2-star structure was missing a classification under 2-hydroxy fatty acid, leading to a flawed program that attempted to find rules to exclude this structure. This missing classification was reported and confirmed by ChEBI curators (https://github.com/ebi-chebi/ChEBI/issues/4767).

Even though the performance of learned programs is lower than deep learning methods when measured over all classes in the benchmark, the two approaches are highly complementary, and potentially can be composed together for a more powerful approach. First, with the program learning approach, we can use the learnability of the class (the train time F1) to filter out classes liable to perform poorly. Second, learned programs can be subsequently refined by curators and cheminformaticians, to hopefully achieve better performance. Third, the deep learning and learned program approaches can be combined in a hybrid system with both increased performance, and the benefits of explainability for the subset of learned classes.

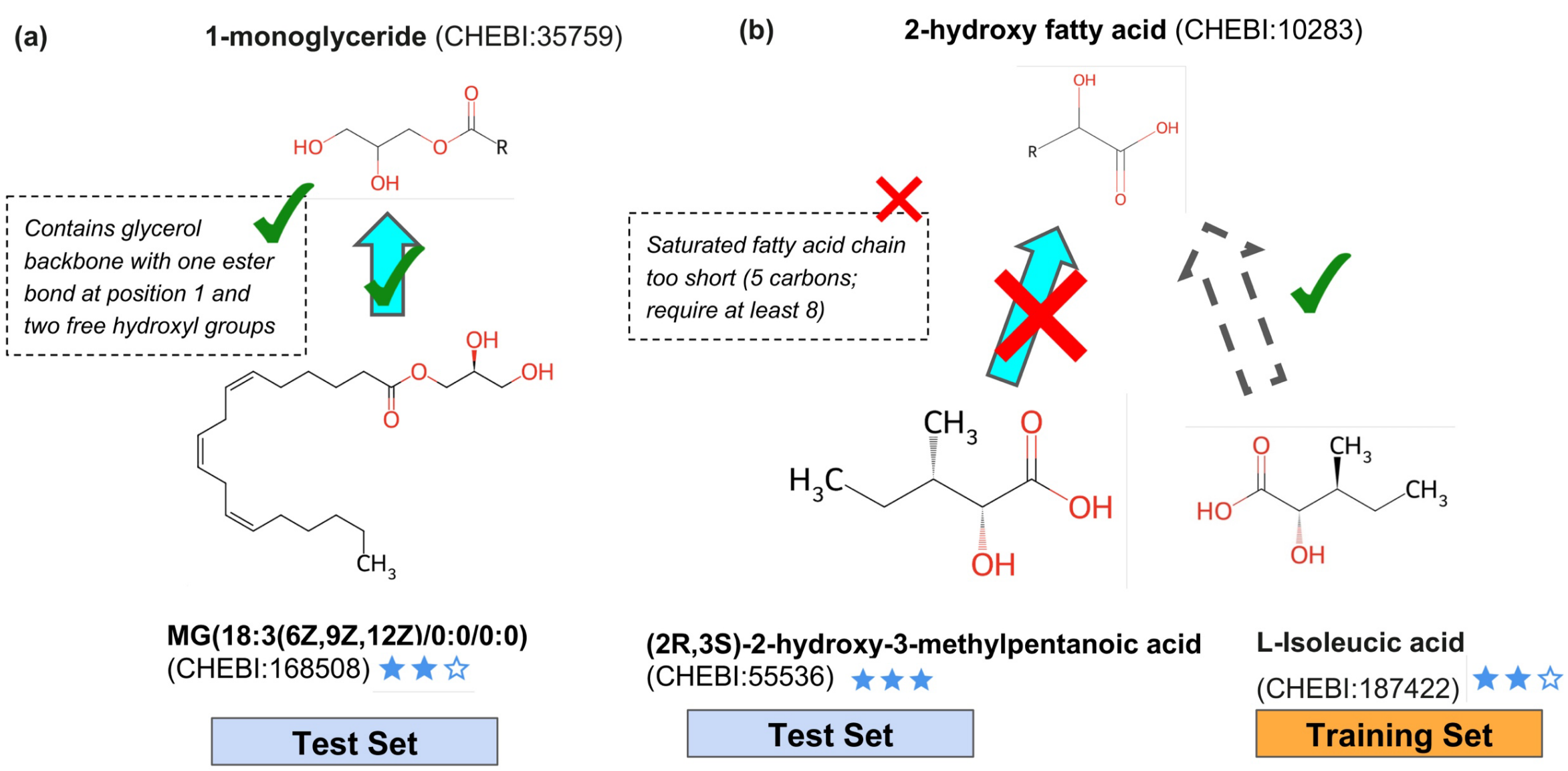

**Figure: 11. Example explanations for both correct and incorrect classifications**
(a) example explanation positive C3PO classification of a ChEBI 2-star structure (MG(18:3(6Z,9Z,12Z)/0:0/0:0)) as a 1-monoclyceride, together with explanation. (b) example of an explanation of an incorrect classification of a ChEBI 3-star structure (2R,3S)-2-hydroxy-3-methylpentanoic acid
) to a 2-hydroxy fatty acid. While the explanation and underlying program was flawed, the explanation helped pinpoint the presence of an missing classification in the training set: L-Isoleucic acid (a ChEBI 2-star structure) should have been classified as a 2-hydroxy fatty acid. The learning trace for this class revealed the program had been overfitted to exclude this structure.

# Discussion

## Limitations

### Errors in imported chemical terms confound the training process and evaluation

As a part of this study, we performed additional checks on the evaluation results, and in particular on cases where learned programs did less well than we expected. For some of the cases that we investigated, we discovered occasional systematic errors that might confound both the learning process and evaluation. An example of this is the generic class ‘cation’, grouping all structures in a positive state. Surprisingly given the simplicity of the class definition, the ensemble of all learned programs did relatively poorly on this, scoring an F1 of 0.78. The deep learning approach has a similar score of 0.81. We expect this class to be trivial for a procedural approach, as there are built in functions in RDKit for calculating charge. We might also expect the deep learning approach to do well, although higher order operations like counting may be harder. However, on examination, we discovered a number of systematic misclassifications based on charge state, and that these led the learning process to produce an overly complex program that overfits the incorrect state (see **Fig. S5**).

When creating the benchmark, we elected to make use of all CHEBI entries, regardless of the ‘star’ rating. In ChEBI, an entry has a 3 star rating if it has been manually checked by the curator team, and a 2 star rating if it is imported from another source (preliminary data has 1 star). There are twice as many 2 star rated terms as 3 star ones (and a much smaller number of 1 star terms), so we elected to use all of these, reasoning that having a larger training set would improve overall performance. In practice this did not turn out to be the case, and the presence of errors of omission and commission in 2 star entries confounded the learning process. An example of this is in the learning of the class ‘2-hydroxy fatty acid’ (CHEBI:10283). The initial programs generated were aligned with the intended meaning of this class, and correctly included structures like ‘L-isoleucic acid’ (CHEBI:187422). However, this structure has not been reviewed by ChEBI curators, reflected in its 2 star rating, and is missing placement under ‘2-hydroxy fatty acid’ in ChEBI, thus was included as a negative example. This means that correct programs receive a lower score. Furthermore, examining the chain of thought process during learning revealed that the LLM had created a complex incorrect heuristic rule to try and rule out this structure as a member of the class, ultimately leading to poorer performance on the test set.

On reflection, the fact that we do not make use of the star rating during the learning process likely leads to poorer performance. One possibility we intend to explore is keeping the 2 star entries during learning, but prompting the model to weight these less strongly if they do not classify in the expected places.

We are currently in the process of performing a more extensive manual evaluation of existing classifications in ChEBI, making use of both Chebifier and C3PO.

## Complexities due to protonation state modeling in ChEBI

For this study we used the C3PO-Slim benchmark, which is derived from ChEBI, filtered based on proxies for biologically meaningful classes. For example, C3PO-Slim contains “amino acid” but not “amino acid zwitterion”, as differentiation between protonation states at the class level is generally not useful from a biological perspective. However, the validation set may indeed include many zwitterion state amino acids, which are formally not classified under “amino acid”. This adds complexity to the generated programs and adds an unnecessary variable to the classification process. This issue was explicitly noted in the original ClassyFire paper[28], and the authors decided to use a simpler ontology that removes such granular distinctions, ultimately leading to a simpler classification task, but reducing the usability of ClassyFire for ChEBI classification.

In the future we aim to investigate the impacts of this complexity, and to create a benchmark based on the simplified Chemessence ontology[52], which collapses protonation state distinctions, making an ontology that is simpler for both humans and machines. We aim to use the approach outlined in this paper to validate the results.

## Naive program learning leads to poor modularity

One downside of our approach is that it lacks modularity – each class has its own standalone program, and there is no reuse of shared components between them. Modularity would improve with overall coherency of the set of programs, and would likely lead to efficiency gains as well. For example, currently there is a major inefficiency in that when classifying across all of C3PO, the SMILES string is repeatedly encoded as an RDKit molecular object, rather than this happening once and then being shared.

An example of where modularity could help is in the above example of learning the program for ‘2-hydroxy fatty acid’, defined in ChEBI as “*any fatty acid with a hydroxy functional group in the alpha- or 2-position*”. This textual definition is inherently compositional, consisting of a genus (fatty acid) and differentia (hydroxy functional group in the alpha- or 2-position). The current learning approach learns a standalone program to classify this structure, rather than reusing the program for fatty acid. Decomposing in this way has advantages for maintainability – for example, if the definition for fatty acid changes, then it only needs to be changed in one place.

Future work could involve using AI to separate out common components into shared modules. This could also be used to automate classification *between* classes, as well as between structures and classes.

### Inherent ambiguities in chemical nomenclature need to be resolved

We found other cases of reduced performance where the meaning of a class was ambiguous. For example, *wax esters* are typically defined to be of length at least 12 (see for example[53]. The top program for classifying wax esters scored 0.94 on the training data, which is good, but short of a perfect score. On further examination, this program includes a constraint that wax esters consist of alkoxy chains of at least 12 carbons in length. ChEBI includes 'decyl palmitate' as a wax ester, but the program rejects this classification with the reason "'*Alkoxy chain too short (10 carbons), need at least 12*".
The synthesized program is arguably correct, in that wax esters are typically defined to be of length at least 12[53]. However, the ChEBI text definition does not mention the length restriction, and 'decyl palmitate' is frequently classified as a wax ester despite its shorter length. Note that the training set did not include any shorter wax esters, which presented the approach with an impossible task, since some of the implicit conventions in ChEBI cannot be fully determined.

Another systematic case of ambiguity is in the classification of biosynthetically derived molecules, and in particular natural products such as terpenoids. In some cases, these might be defined structurally, and in other cases biosynthetically, in terms of the pathway that produced them. This ambiguity leads to a fuzzy classification, where for any one class, some structures were curated one way, and others in another way. Deriving programs for true biosynthetically defined classes is impossible using structure alone, and this would require reasoning over biochemical pathway databases such as KEGG, GO, or BioCyc.

These ambiguities confound both the learning process and evaluation. In future, both AI and manual curation will be required to tease apart these ambiguities.

## Future Directions

### Applications for automating ChEBI classification

We demonstrated that chemical classification programs can be learned using an iterative code generation approach with accuracy close to state of the art deep learning methods. This has immediate applications for helping streamline the internal ChEBI curation process, and for automatically classifying chemical structures in chemical databases using ChEBI.

One challenge with our approach is that our method is sensitive to errors in the training set. As we have demonstrated, errors of omission and commission in the source ontology can lead to the generation of programs that overfit in order to recapitulate incorrect classifications. Manually re-curating the training set would be highly time consuming and expensive. We could potentially combine the existing ChEBI ontology, ClassyFire, and Chebifier and select only classifications agreed on by all three to create a more reliable set. However, the three methods are not independent, and if one has a mistake it is more likely the others do (because ChEBI uses ClassyFire in a semi-automated fashion, and Chebifier is trained on ChEBI).

An alternative is to use an incremental process, similar to how OWL definitions and OWL reasoning is used in many OBO ontologies; definitions need not be complete for them to be useful. We would start by taking the highest scoring chemical class programs and using these as part of the process for assigning chemical structure parents. The results would of course be examined by ChEBI curators; if the provided classification conflicts with either the chemical knowledge of the curator, or the results of using a different automated classification system (e.g. Chebifier or ClassyFire) then this would be investigated further, and the program could be adjusted. This involves some initial investment of effort, but as confidence is gained in the system, then programs could be marked as being trusted, and used purely in an automated fashion (this could even be done ahead of time, through auditing the program). The programs could be managed in GitHub, with the broader community making contributions via pull requests.

This process could be accelerated even further, by the use of newer agent-based AI methods.

## Using agentic AI to simultaneously learn programs and repair training set errors

In this work, we intentionally used a predefined workflow, in which we attempted to optimize generated programs to maximize scoring on the provided benchmarks. This allowed us to evaluate the impact of model choice, hyperparameters, and relative performance on different kinds of chemical structures. However, the methods are sensitive to errors in training sets, and do not allow the AI to explore a more open-ended approach.

One possibility is to use agent-based AI methods to simultaneously learn programs in combination with exploring the literature and trusted sources such as IUPAC. We have started exploring the use of agentic AI for ontology development[54], and plan to extend this to chemical classification.

## Creating hybrid classifiers from complementary approaches

In this study, we showed the complementary nature of learned program based classification, and the deep learning approach exemplified by Chebifier. Combining these and other approaches, including logic based classification, could provide a means to yield both more accurate classifications, and to enhance existing methods with explanations. We have embarked on an investigation of hybrid strategies, and have implemented a preliminary version of this in the Chebifier 2 command line tool and web-based tool[55].

## Applications of classifier programs to other domains

For this study we generated Python programs for binary chemical structure classifiers, using the RDKit library. In theory, the general approach of learning classifier programs could work on any domain. This procedure could also be used as a form of knowledge distillation, taking black box classifiers, and learning rules that explain their behavior. In practice, the approach is likely better suited to complex domains involving complex interconnected entities, making use of specialized libraries such as RDKit.

In the future, we aim to test the LEIA approach on related domains such as learning classification rules for enzymatic reactions, such as those represented in the RHEA database[56], as well as for learning rules for biosynthetic gene clusters (BGCs).

# Conclusions

Accurate and scalable classification of chemical structures into classes is crucial for multiple applications in biology, health, and environmental science. Using deep learning to directly classify structures can have high accuracy, but suffers from lack of explainability, interpretability, and the lack of the ability for curators to directly manipulate models. Logic and rule based approaches require significant manual effort to curate definitions, and lack the higher level chemical abstractions (e.g. counting rings) used by chemists when reasoning about classification.

We have demonstrated the feasibility of generating a classifier program ontology for chemical structures using generative AI learning techniques. The resulting programs are explainable, execute deterministically, and provide explanations for classifications. Although these currently do not yet reach the performance on state of the art deep learning models, they provide a complementary approach and could be directly incorporated into curation workflows, where they could be improved as part of an iterative process.

# List of abbreviations

- AI: Artificial Intelligence
- C3PO: ChEBI Chemical Classification Program Ontology
- ChEBI: Chemical Entities of Biological Interest
- LEIA: Learn-Execute-Iterate-Adapt
- LLM: Large Language Model

# Declarations

## Availability of data and materials

All code for learning, evaluation, and analysis is available in the c3p github repository: https://github.com/chemkg/c3p. The C3PO Benchmark has been deposited on Hugging Face (http://doi.org/doi:10.57967/hf/4033, https://huggingface.co/datasets/MonarchInit/C3PO).

An accompanying website documenting this is available at https://chemkg.github.io/c3p.

## Competing interests

None

## Funding

CJM and JR were supported by the Genomic Science Program in the U.S. Department of Energy, Office of Science, Office of Biological and Environmental Research (BER) under contract numbers DE-AC02–05CH11231. CJM was also funded by NIH HG012212. CJM and DK were funded in part by NIH HG010860. JH acknowledges funding from the Swiss National Science Foundation under grant agreement numbers 215906 (StrOntEx) and 10002786 (MetaboLinkAI). AM and NMOB acknowledge funding from the European Molecular Biology Laboratory.

## Authors Contributions

CJM devised the approach and ran the initial experiments. AM performed curation and led the evaluation of LLM results. JH and NMOB advised on the strategy and provided evaluation use cases, and contributed to the evaluation. JR and DK assisted with the analysis. All authors contributed to the writing and review of the manuscript.

## Acknowledgements

We thank Andrew Schmeder and the Berkeley Lab CBORG team (https://cborg.lbl.gov/) for providing access to multiple models for evaluation purposes, and Nomi Harris and Melissa Haendel for comments and guidance on the manuscript.

# Supplementary Figures

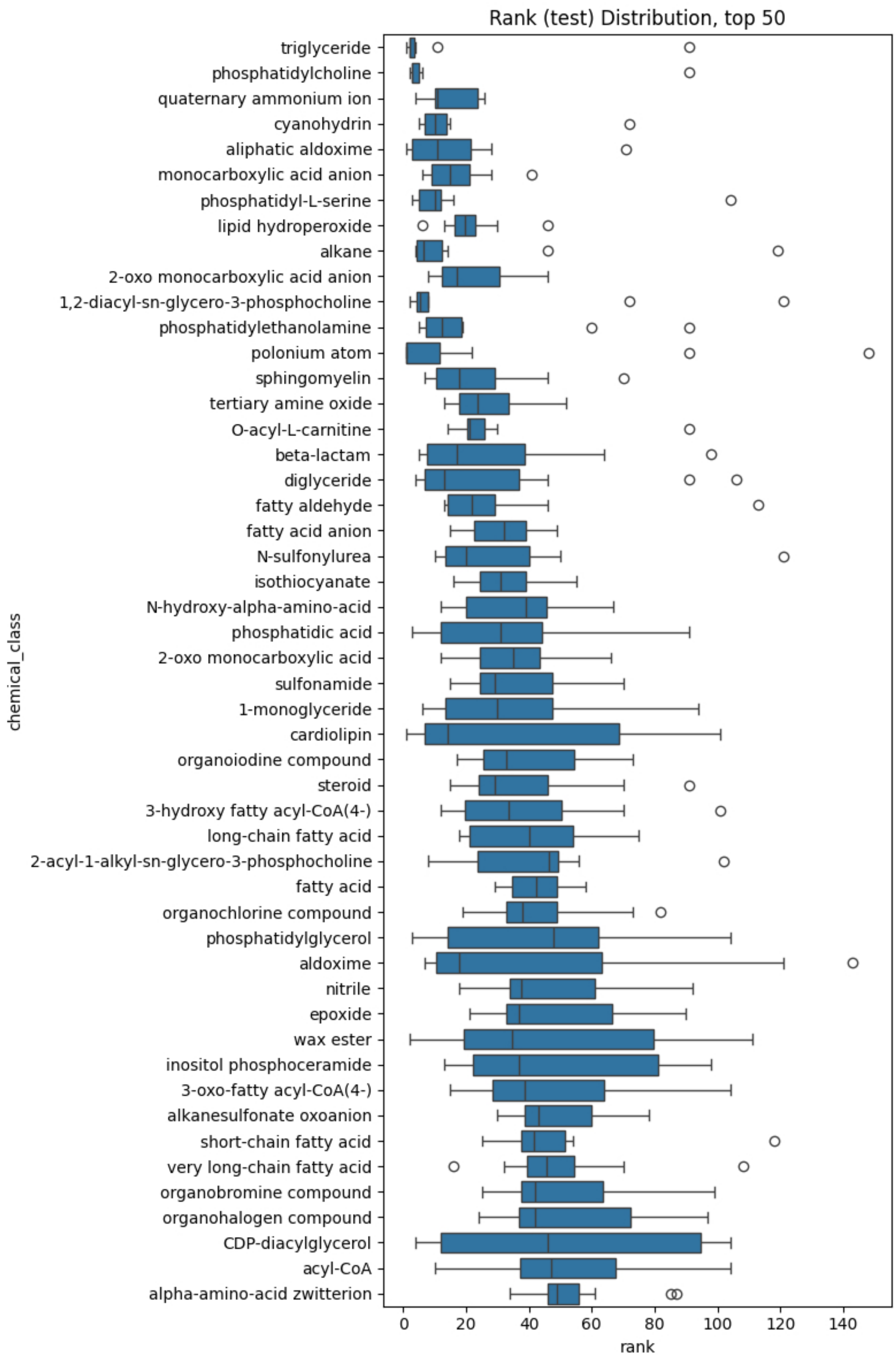

Rank (test) Distribution, top 50
chemical_class
rank
triglyceride
phosphatidylcholine
quaternary ammonium ion
cyanohydrin
aliphatic aldoxime
monocarboxylic acid anion
phosphatidyl-L-serine
lipid hydroperoxide
alkane
2-oxo monocarboxylic acid anion
1,2-diacyl-sn-glycero-3-phosphocholine
phosphatidylethanolamine
polonium atom
sphingomyelin
tertiary amine oxide
O-acyl-L-carnitine
beta-lactam
diglyceride
fatty aldehyde
fatty acid anion
N-sulfonylurea
isothiocyanate
N-hydroxy-alpha-amino-acid
phosphatidic acid
2-oxo monocarboxylic acid
sulfonamide
1-monoglyceride
cardiolipin
organoiodine compound
steroid
3-hydroxy fatty acyl-CoA(4-)
long-chain fatty acid
2-acyl-1-alkyl-sn-glycero-3-phosphocholine
fatty acid
organochlorine compound
phosphatidylglycerol
aldoxime
nitrile
epoxide
wax ester
inositol phosphoceramide
3-oxo-fatty acyl-CoA(4-)
alkanesulfonate oxoanion
short-chain fatty acid
very long-chain fatty acid
organobromine compound
organohalogen compound
CDP-diacylglycerol
acyl-CoA
alpha-amino-acid zwitterion
0
20
40
60
80
100
120
140

## Figure S1: Distribution of ranks for average top learnable classes

Each row is a chemical class, with the plot showing the distribution of relative rankings for each model. Classes such as trigylceride are ranked as among the best scoring broadly across all experiments, but in some outlier cases this ranked more poorly. Note that Figure 4 shows the most learnable classes for the ensemble, whereas this gives a picture of variation among models and hyperparameters.

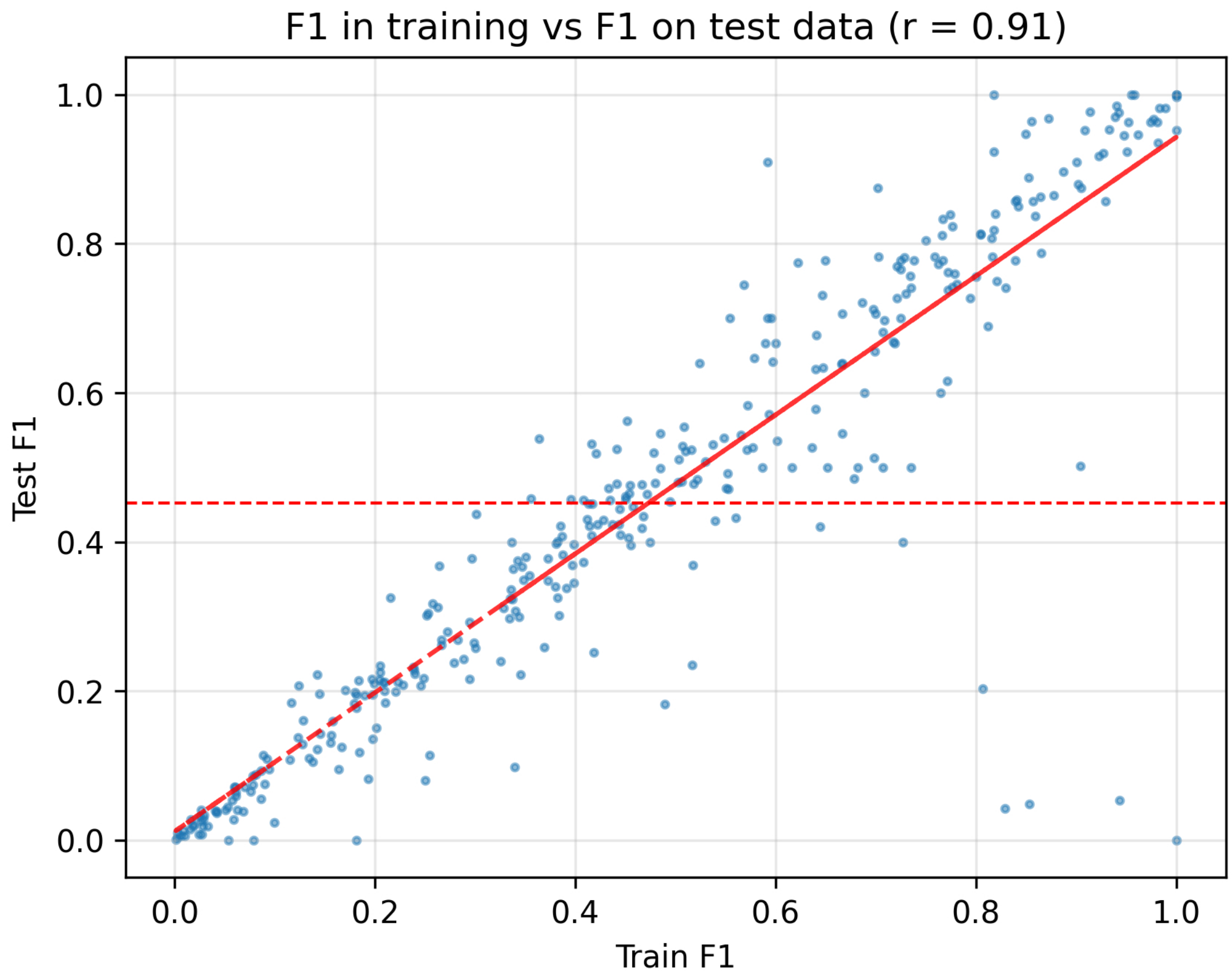

## Figure S2: Generalization from training to test F1 scores.

Scatterplot of F1 score from training phase vs F1 score from held-out data in test phase, using ensemble model. There is a high correlation, with a broad distribution of scores. This indicates we can know in advance with some confidence which classifications will be reliable, i.e. more learnable classes predictably perform better.

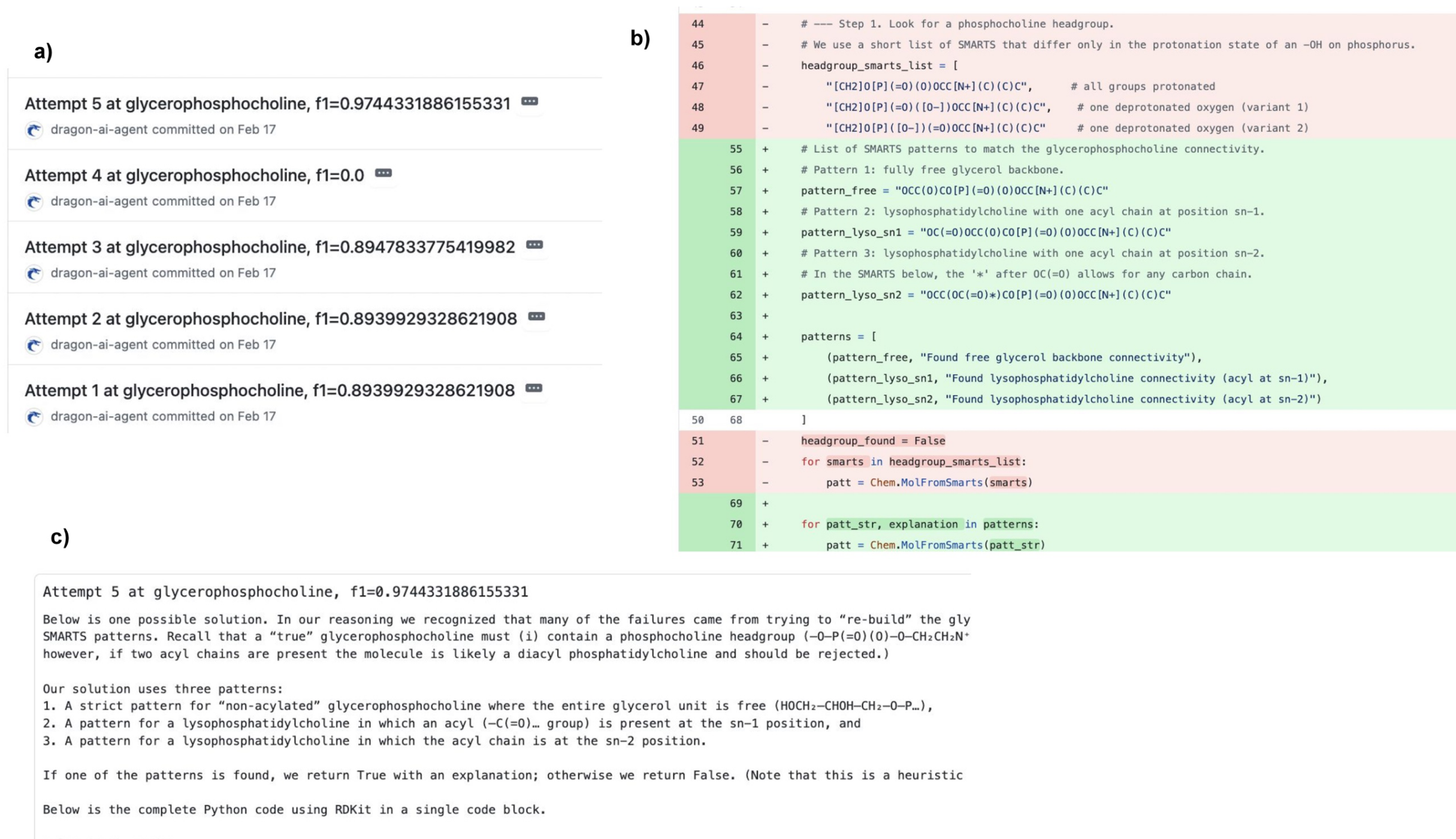

# Figure S3 Example of program convergence, for glycerophosphocholine

(a) GitHub log of attempts made by the agent to produce the best program (using only training examples), with progression from 0.89 to 0.97, with a temporary ‘blind alley’ exploration. (b) Example of one step program evolution, showing code difference between 4th and 5th attempt (c) agent’s rationale for the changes made to create the final (5th) attempt (stored as comments in GitHub).

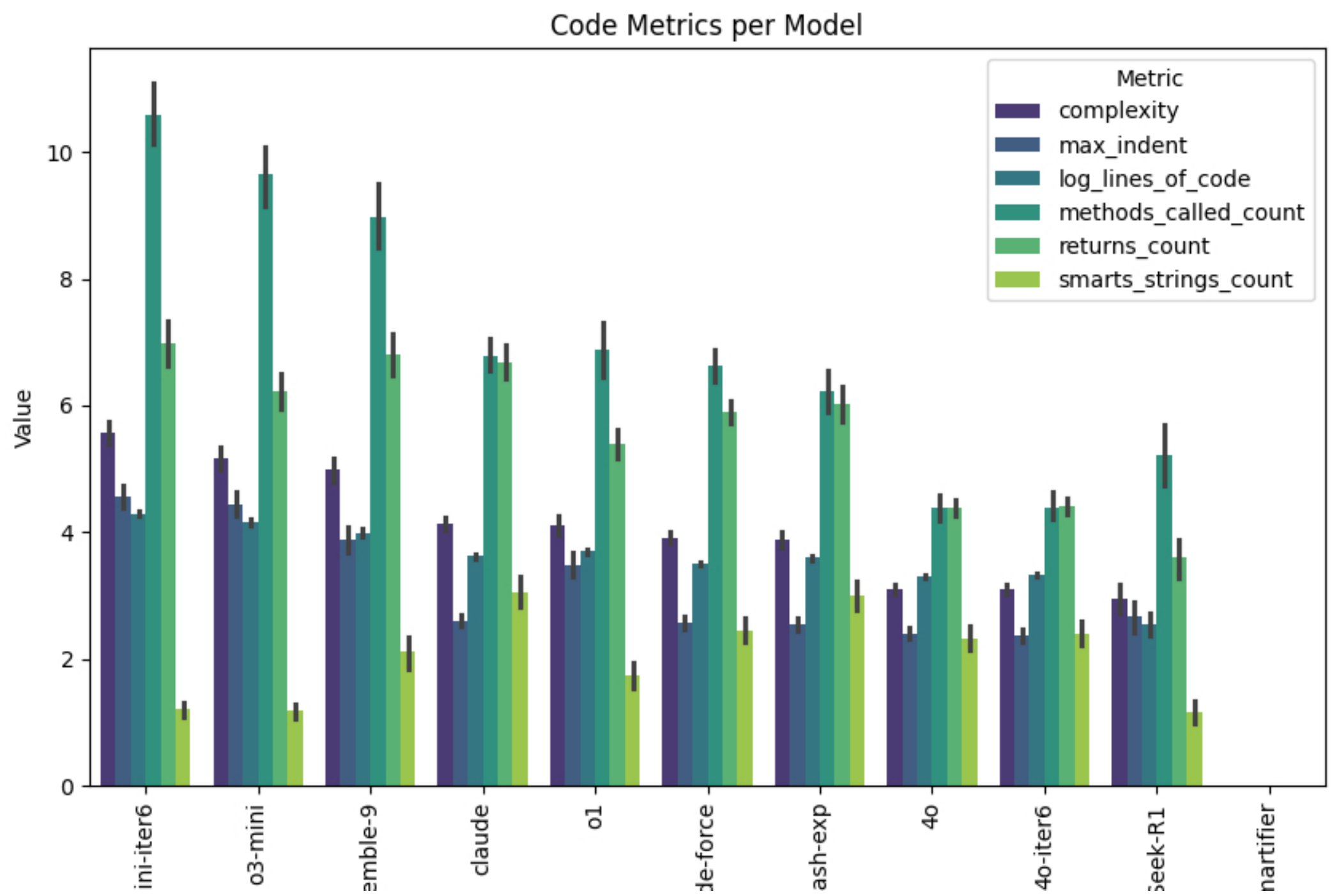

# Figure S4 Analysis of code properties of generated programs

Different metrics include total lines of code (log scale), number of distinct methods called, number of return points (as a proxy for different branch points in decision logic), the number of different SMARTS strings used. Error bars indicate 95% confidence. Overall o3 mini produced the most complex code, and gpt-4o (in all 3 configurations) was the most laconic, favoring shorter, less complex programs. Different models varied widely in how they used SMARTS strings, with Claude averaging the most, and o1 using them the least.

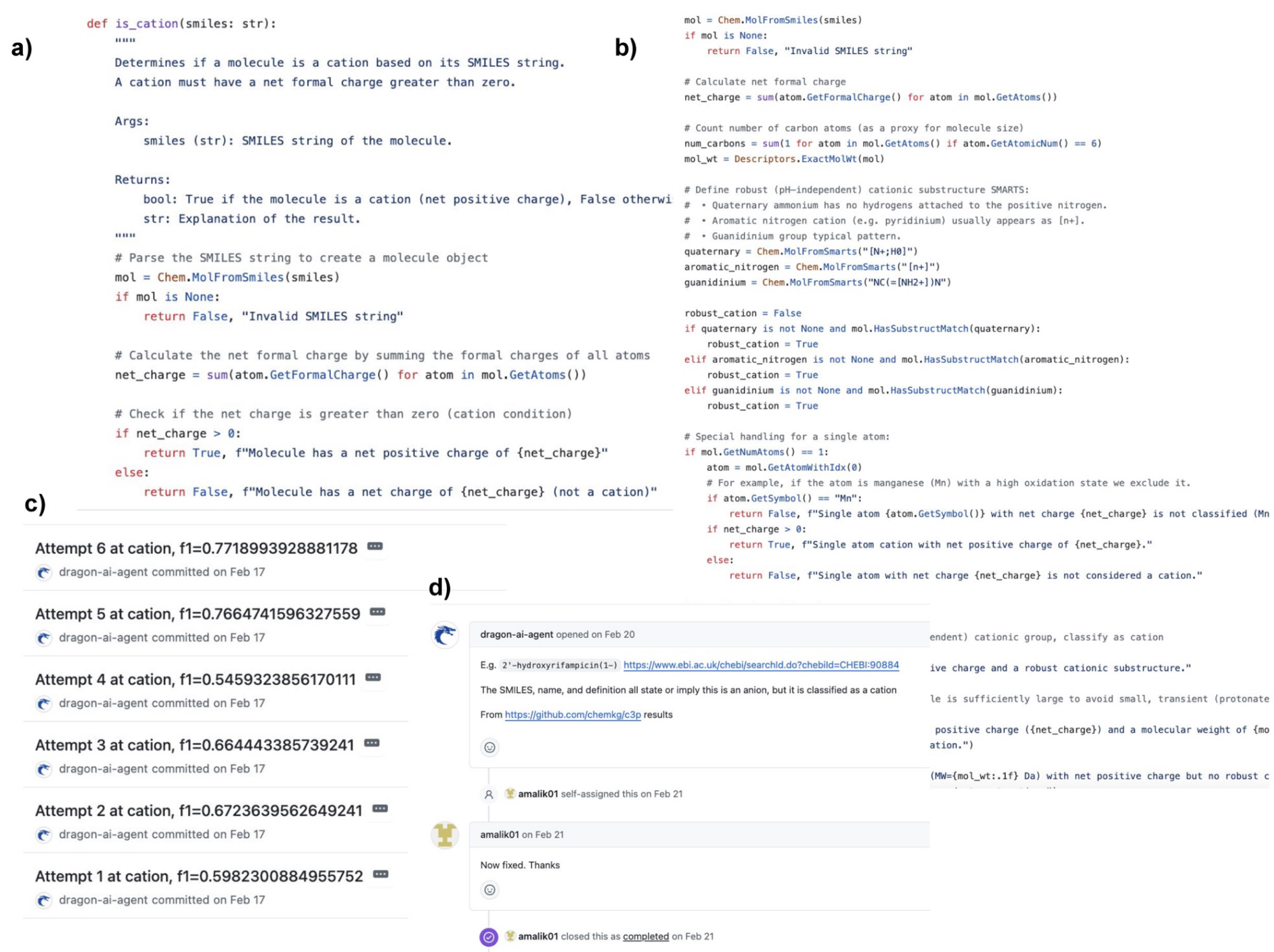

## Figure S5 Example of a case where errors in the training set cause learning to overfit on overly complex programs

(a) the initial first pass at a program is parsimonious and logically valid, yet this had F1=0.6. (b) the final program converged on after five subsequent attempts. This is considerably more complex, but it achieves a higher F1 score (c) evolution of F1 scores. (d) an example of an issue created in the ChEBI tracker highlighting the mistaken classification and resolution (in fact there were many other such misclassifications). Not shown is the thinking process for generating the final version of the program, repeated below:

*“Below is our “thinking-out-loud” summary before the code. In our last attempt the logic was very “binary” (simply net charge positive versus zero) so that many small, protonated species got classified as cations even though their positive charge is pH–dependent, while many zwitterionic molecules with a robust “cationic lipid” character (for example, many phosphatidylcholines) were missed. To improve the F1 score we (i) now distinguish “robust” cationic groups (i.e. quaternary ammonium, aromatic nitrogen cations, guanidinium groups) from protonated amines; (ii) we also require that—for molecules in the net–zero case—the compound must contain a phosphorus atom (as found in a phospholipid headgroup) and have a large carbon skeleton (we now require at least 30 carbon atoms) in order to call it a cation (this helps avoid mis-calling small zwitterions as “cationic”); and (iii) for singly charged molecules whose positive charge comes from a*

*protonated amine (i.e. not “robust”) we now impose a molecular-weight threshold (approximately 200 Da) so that very small species (such as N–dimethylethanolamine, MW≈90) are not called cations. In addition, when a molecule is a single atom we accept it as a cation—except in one ad hoc case (for example, manganese in the unusual +7 oxidation state) which we explicitly exclude.”*

# Supplementary Tables

## Table S1: Derived molecular descriptors

## Table S2: Full results for C3PO (ensemble) and Chebifier